\pdfoutput=1

\documentclass[11pt]{article}

\usepackage{coling}
\usepackage{times}
\usepackage{latexsym}
\usepackage{graphicx}
\usepackage{tabularx}
\usepackage{subfigure}
\usepackage{mathrsfs}
\usepackage{amsthm,amsmath,amssymb}
\usepackage{mathrsfs}
\usepackage{amsmath}
\usepackage{tcolorbox}
\usepackage{color}
\usepackage{amssymb}
\usepackage{makecell}
\usepackage{array}
\usepackage{times}
\usepackage{latexsym}
\usepackage{listings}
\usepackage{multirow}
\usepackage{array}
\usepackage[T1]{fontenc}

\usepackage[utf8]{inputenc}

\usepackage{microtype}
\usepackage{booktabs}
\usepackage{multirow}
\usepackage{booktabs}
\usepackage{amssymb}
\usepackage{graphicx}
\usepackage{times}
\usepackage{latexsym}
\usepackage{url}
\usepackage[normalem]{ulem}
\useunder{\uline}{\ul}{}
\usepackage{inconsolata}

\usepackage{graphicx}

\title{
\textit{Sibyl}: Empowering Empathetic Dialogue Generation in Large Language Models via Sensible and Visionary Commonsense Inference
}

\author{Lanrui Wang\textsuperscript{\rm 1,2}\thanks{\ \ \ Equal contribution. }, Jiangnan Li\textsuperscript{\rm 3}\footnotemark[1],  Chenxu Yang\textsuperscript{\rm 1,2}, Zheng Lin\textsuperscript{\rm 1,2}\thanks{\ \ \ Zheng Lin is the corresponding author. },\\ {\bf Hongyin Tang\textsuperscript{\rm 4},
 Huan Liu\textsuperscript{\rm 1}, Yanan Cao\textsuperscript{\rm 1,2}, Jingang Wang\textsuperscript{\rm 4}, Weiping Wang\textsuperscript{\rm 1}}  \\
  \textsuperscript{\rm 1}Institute of Information Engineering, Chinese Academy of Sciences, Beijing, China \\
  \textsuperscript{\rm 2}School of Cyber Security, University of Chinese Academy of Sciences, Beijing, China \\
  \textsuperscript{\rm 3}WeChat AI, Tencent Inc, China \quad \textsuperscript{\rm 4}Meituan Group, Beijing, China  \\
  \texttt{\textrm{\{}wanglanrui,yangchenxu,linzheng,wangweiping\textrm{\}}@iie.ac.cn} \\
  \texttt{jiangnanli@tencent.com}  \quad \texttt{\textrm{\{}tanghongyin,wangjingang02\textrm{\}}@meituan.com}
  \\
  }
  
\begin{document}
\maketitle
\begin{abstract}
Recently, there has been a heightened interest in building chatbots based on Large Language Models (LLMs) to emulate human-like qualities in multi-turn conversations. Despite having access to commonsense knowledge to better understand the psychological aspects and causality of dialogue context, even these powerful LLMs struggle to achieve the goals of empathy and emotional support. 
Current commonsense knowledge derived from dialogue contexts is inherently limited and often fails to adequately anticipate the future course of a dialogue. This lack of foresight can mislead LLMs and hinder their ability to provide effective support. 
In response to this challenge, we present an innovative framework named \underline{S}ens\underline{ib}le and Visionar\underline{y} Commonsense Know\underline{l}edge (\textit{Sibyl}). Designed to concentrate on the immediately succeeding dialogue, this paradigm equips LLMs with the capability to uncover the implicit requirements of the conversation, aiming to elicit more empathetic responses.
Experimental results demonstrate that incorporating our paradigm for acquiring commonsense knowledge into LLMs comprehensively enhances the quality of their responses\footnote{\ The code is available at \url{https://github.com/wlr737/Sibyl}}.

\end{abstract}

\section{Introduction}

Empathy, in its most comprehensive definition, is the reaction of one individual to the observed experiences of another \citep{davis1983measuring}. 
Given the inherent complexity of conversation, recent works focus on integrating commonsense knowledge to aid in unraveling the implicit psychological motivations and causality within utterances \citep{ MISC, EmpSoA}. Meanwhile, sophisticated abilities of Large Language Models (LLMs) in dialogue understanding and response generation have ignited a new zeitgeist for building a powerful dialogue agent \citep{ChatGPT, GPT4, Llama2, llama3}. 
By incorporating commonsense knowledge as reasoning steps \citep{Cue-CoT, DialogueCoT}, which prompt additional vital associative information in dialogue contexts, these LLMs demonstrate human-like abilities in understanding and generating dialogue responses.


\begin{figure}
    \centering
    \includegraphics[width=0.48\textwidth]{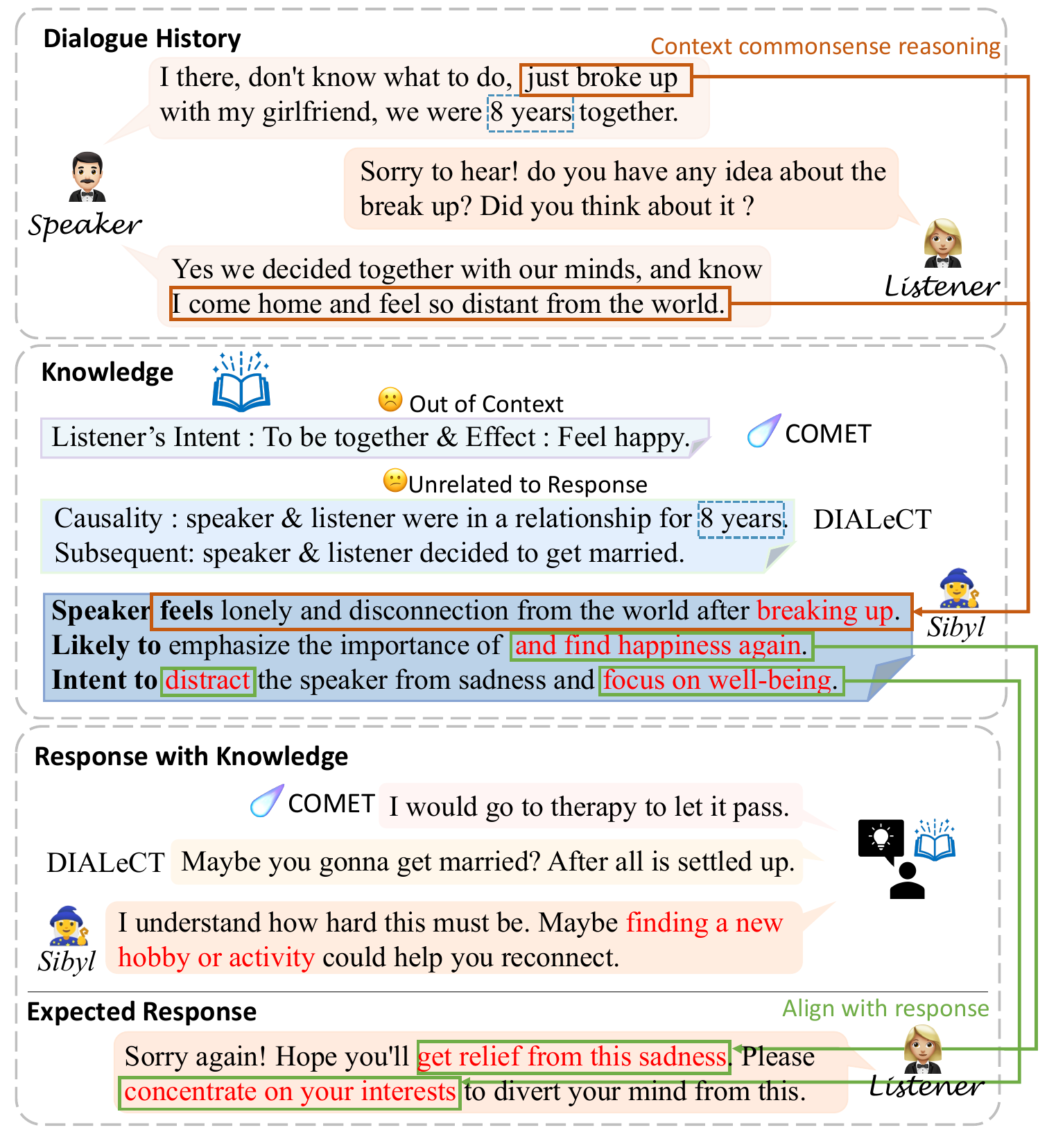}
    \caption{An example from the \textsc{EmpatheticDialogues} dataset reveals that the commonsense inference deduced by COMET and DIALeCT demonstrates notable limitations.} 
    \label{fig: cases}
\end{figure}

Despite their notable successes, these advanced LLMs still struggle to produce empathetic responses and provide emotional support in multi-turn conversations \citep{IsChatGPT}.
As illustrated in Figure \ref{fig: cases}, the commonsense inference derived from COMET \citep{COMET} primarily concentrates on the last utterance of the Speaker. This narrow focus fails to correspond with the full context of the multi-turn conversation and inaccurately captures the Speaker's emotional state, leading to cascade errors in generating responses.
Meanwhile, DIALeCT \citep{CICEROv2} employs commonsense reasoning for a complete and static dialogue. This limitation increases the risk of inaccuracies, stemming from its sole focus on dialogue history. The commonsense knowledge deduced by DIAleCT was often unrelated to the response and even misinterpreted the background information of the participants.

Investigating the above phenomenon, we suggest that the issue arises since \textbf{current approaches do not adequately anticipate dialogue future}. 
Due to the one-to-many nature of dialogue generation, basic commonsense knowledge derived from dialogue contexts is inherently restricted. 
The existence of multiple distinct responses that can appropriately answer the same dialogue history suggests that within a given context, there are diverse dialogue commonsense inferences associated with each possible response \citep{ProphetChat, Reflect}.
Exclusively deduced from dialogue history, 
contemporary methods are prone to introducing noisy information and confusing language models to ignore the demand for empathy and emotional support. 





In response to these challenges, this paper presents a new paradigm that dynamically deduces commonsense knowledge relevant to the prospective future of dialogue, called \underline{S}ens\underline{ib}le and Visionar\underline{y} Commonsense Know\underline{l}edge (\textit{Sibyl}).
This involves instructing models to identify potential causal factors from the prior dialogue history, along with the mental states of participants and their possible intentions for upcoming statements. 
We argue that\textbf{ the dialogue history does not encompass enough information to generate the intended response.} By deriving plausible future-aware commonsense knowledge from prophetic powerful LLMs, we empower open-source LLMs to generate these visionary inferences solely based on dialogue history. Essentially, these visionary inferences act as a form of chain-of-thought (CoT) prompts, aiding LLMs in effectively dealing with complex dialogue contexts, \textbf{bridging the gap between dialogue history and potential response}, and ultimately promoting empathy and emotional support. They furnish crucial implicit information regarding emotional states, intentions, subsequent events, and the scope of dialogue context that can elicit the desired response in the conversation. In-depth experiments on the EmpatheticDialogues~\cite{EDdata} and Emotional Support Conversation~\citep{ESCdata} datasets demonstrate the superiority of \textit{Sibyl} over competitive categories of commonsense knowledge when applied to LLMs under multiple settings.

In summary, our contributions are as follows:
 \begin{itemize}
    \item Our research addresses the shortcomings of current commonsense inference methods in anticipating dialogue future, arising from the one-to-many issue. Multiple pieces of commonsense knowledge linked to a single context can confuse language models, causing them to ignore the goals of fostering empathy and providing emotional support.
    \item We propose \textit{Sibyl}, an innovative paradigm for multi-turn dialogue commonsense inference that encompasses psychological, emotional, and causality factors in commonsense inference, which is pertinent to dialogue future. 
    \item Extensive experiments demonstrate the effectiveness of our paradigm, and detailed analyses validate our method across multiple scenarios and backbone models, showing significant improvements in automated metrics and evaluations by human and powerful LLM assessors.
\end{itemize}

\begin{figure*}
    \centering
    \includegraphics[width=1\textwidth]{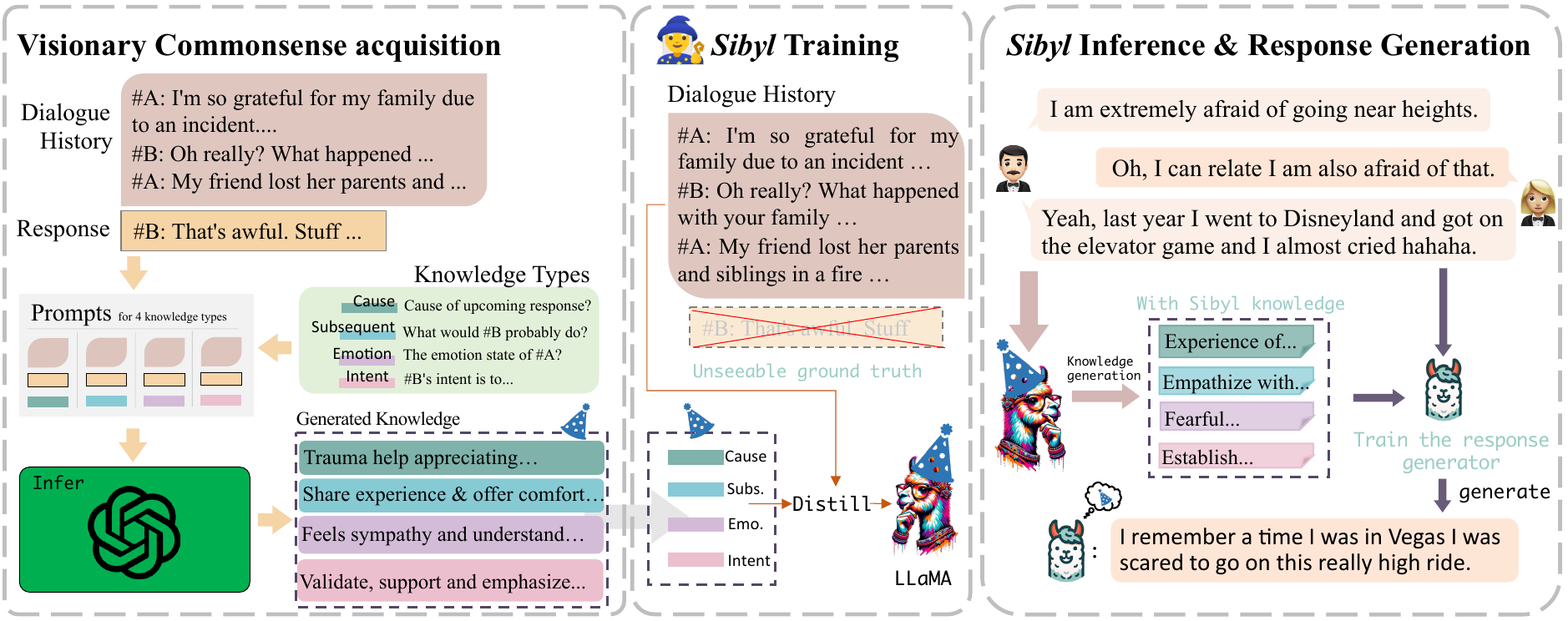}
    \caption{
    The overview of our proposed paradigm of Commonsense Inference, \textit{Sibyl}. Incorporating both dialogue history and ground truth responses, the powerful LLM first deduces four categories of visionary commonsense. These inferences serve as a guiding oracle, aiding LLaMA models in inferring from dialogue history alone during the training stage. Subsequently, these trained models function as experts in inferring four categories of commonsense knowledge.
    }
    \label{fig: model}
\end{figure*}

\section{Related Work}

\textbf{Empathy} refers to the capacity to anticipate and understand the reactions of others \citep{keskin2014isn}. Early studies concentrated on producing empathetic dialogues by leveraging the Speaker's emotional signals \citep{MoEL, MIME} within the \textsc{EmpatheticDialogues} dataset \citep{EDdata}. To enhance the ability to understand, perceive, and respond appropriately to the situation and feelings of others, commonsense knowledge is widely incorporated into empathetic chatbots \citep{CEM, KEMP, SEEK, CASE}. Recently, several research efforts have explored the application of LLMs in generating empathetic responses within a prompt-based framework revealing the limitations of LLMs in accomplishing this task \citep{IsChatGPT, ED_ChatGPT}.

Empathy has also been related to several other variables such as helping, introversion, and affiliative tendency \citep{chlopan1985empathy}. \textbf{Emotional Support Conversation} \citep{ESCdata} is a benchmark focusing on exploring the problem of help seekers and generating more supportive responses. COMET \citep{COMET}, a pre-trained generative commonsense reasoning model is employed to obtain commonsense knowledge of the dialogue \citep{MISC, GLHG, KEMI}. However,  in the absence of harmonious knowledge selection, external information might trigger logical conflicts in dialogue \citep{TAKE, SEEK}.


\textbf{Commonsense knowledge} plays a vital role in dialogue systems, with numerous studies focusing on improving its acquisition techniques. \citet{COMET} offers local utterance-wide commonsense inference, widely utilized in dialogue systems. \citet{CICEROv1, CICEROv2} train language models to produce context-aware commonsense knowledge through natural language generation (NLG) and multiple-choice question (MCQ) tasks, advancing the application of commonsense knowledge in dialogue for further research. Recent studies indicate that commonsense reasoning \citep{DiaCoT_and_PPO, Simbolic}, derived through \textbf{multi-step} methodologies that function like chain-of-thought prompting, markedly outperforms the approach of prompting LLMs to concurrently deduce implicit information and generate dialogue responses \citep{Cue-CoT, Frugal_Prompting}. By appending commonsense knowledge to the dialogue context \citep{Cue-CoT, DialogueCoT}, these inferences serve as intermediate reasoning to trigger LLM analysis and produce high-quality responses.

\section{Preliminaries}

\subsection{Problem Formulation}
In the task of dialogue response generation, we employ $\theta$ to signify a dialogue model, while $C=[u_1, u_2, ..., u_{n-1}]$ indicates the context utterances, and $K$ corresponds to commonsense knowledge. The objective here is to predict the forthcoming response $Y$  based on the given context $C$ from the $n-1$ turn, supplemented with the external commonsense knowledge $K$. 

\begin{equation}
Y \sim P_\theta(\cdot\mid K,C)
\label{eq: 1}
\end{equation}

\subsection{Categories of Commonsense Inference}
This study incorporates four categories of commonsense inferences within dialogues, which include: 1) \textbf{Cause}: Identifying the possible cause in the dialogue history for the forthcoming response. 2) \textbf{Subsequent Event}: Events that might take place in the succeeding dialogue. 3) \textbf{Emotion state}: The user's emotional state as indicated in their latest utterance. 4) \textbf{Intention}: The probable dialogue intent behind the assistant's next response. The overarching goal is to enhance the understanding of dialogue history and meticulously project potential traits of the possible upcoming responses. These inferences operate as crucial intermediate reasoning steps that assist language models in enhancing dialogue comprehension and producing empathetic and supportive responses, with further details in Appendix \ref{appendix_cateies_cmk}.

\section{Method} 
In this section, we propose a novel paradigm for obtaining visionary commonsense knowledge, named \textit{Sibyl}, as demonstrated in Figure \ref{fig: model}.

\subsection{Visionary Commonsense Acquisition} \label{Prophet_acqu}
The advanced LLMs which are aligned with human intention, exhibit robust logical deduction abilities. Initially, we utilize GPT-4o to generate four categories of commonsense inferences $\mathcal{K}$, using inputs that include dialogue history $\mathcal{C}$ and the response $\mathcal{Y}$. 
 We randomly selected a sample as demonstration to guide the powerful LLM in generating a visionary commonsense inference, considering dialogue history and response\footnote{To prevent information leakage, all dialogue samples mentioned in this section are sourced exclusively from the training sets.}.

\begin{equation}
\mathcal{K} =  \underset{K}{{\arg\max} \, P_{\textit{LLM}}} (\mathcal{C} ; \mathcal{Y})
\label{eq: 2}
\end{equation}

To confirm the reasonableness of the four knowledge categories, we employ five human-sourced professional annotators to perform a binary evaluation on 400 randomly chosen samples of commonsense knowledge. The average scores for the knowledge categories all exceed \textbf{0.89}\footnote{The human annotators recruited for this evaluation are the same as those mentioned in Section \ref{human_eval}. The Fleiss's \textbf{Kappa} measure among annotators stands at 0.52, signifying a moderate level of agreement.}.  

\subsection{\textit{Sibyl} Training} \label{Prophet_train}
To independently generate visionary commonsense inferences based on dialogue history, we further undertake Supervised Finetuning (SFT) of open-source LLMs to learn how to cultivate their prophetic abilities. Notably, differing from the process outlined in Sec. \ref{Prophet_acqu}, these aspect-specialized models are presented with input encompassing \textbf{solely the dialogue history}. In other words, they are trained to anticipate the imminent dialogue future, under the instruction of powerful LLMs which possess prior knowledge about the possible response.
Denoted as $\Psi$, these visionary models are trained for analyzing causality, psychology, subsequency, and intent aspects of unseen conversations. 

Prompts templates are carefully designed as hints to guide these models to understand the purpose of performing commonsense inference. Similar to prompting LLMs to generate oracle commonsense inference, we describe the aim of deducing a certain aspect of commonsense knowledge first and give one example of dialogue for LLMs to grasp the demand of reasoning implicitly. The details of the prompt templates discussed in Sections \ref{Prophet_acqu} and \ref{Prophet_train} are illustrated in Figures \ref{figure_visionary_commonsense_acquisition} and \ref{figure_sibyl_training}.

The training loss is the standard negative log-likelihood (NLL) loss on the commonsense knowledge inferred by LLMs:
\begin{equation}
\mathcal{L}_{Infer} = -\sum_{m=1}^{M}log(P(k_m|\mathcal{C},k_{<m}))
\label{eq: 3}
\end{equation}
where $M$ is the length of commonsense inference generated by powerful LLMs, $\mathcal{K}=[k_1,...,k_M]$. 

\subsection{\textit{Sibyl} Inference and Response Generation}
After the training phase of visionary language models, we apply these well-trained visionary models to deduce the mentioned four categories of commonsense knowledge focusing on dialogue future. 
In practice, we take the prompt $C_p$ as the input of models $\Psi$, and we obtain four types of visionary commonsense inference $\mathcal{K}_p$.
\begin{equation}
C_p = Prompt_{template}(\mathcal{C})
\label{eq: 4}
\end{equation}
\begin{equation}
\mathcal{K}_p = \Psi(C_p)
\label{eq: 5}
\end{equation}
where $\mathcal{C}$ indicates dialogue context, the prompt template is detailed in Figure \ref{figure_sibyl_training}, which is consistent with the template used in the training stage, as mentioned in Sec. \ref{Prophet_train}.\\
\textbf{Response Generation.} For response generation, we append all four categories of visionary commonsense inferences $\mathcal{K}_p$  to the corresponding context to compose the input of LLMs. These inferences act as a bridge between dialogue history and the next response, aiding the foundation models to envision the future based on these cues for the probable response. 

We conduct experiments using two strategies for creating the response generator: a finetuned approach and a prompt-based approach using LLMs. The finetuned approach involves two prominent open-source models: LLaMA3.1-8B-Instruct \citep{Llama3.1}, and \textit{Flan-t5-xl} \citep{flant5}.  Standard NLL loss is adopted for the ground truth response $Y$ during the finetuning process:
\begin{equation}
\mathcal{L}_{gen} = -\sum_{g=1}^{G}log(P(y_g|C;\mathcal{K}_p,y_{<g}))
\label{eq: 6}
\end{equation}
where G stands for the length of the ground truth response of the dialogue, $y_g$ specifies the $g$-th token in target response $Y$.

In the prompt-based approach, we directly engage an LLM to generate the subsequent response. The prompt provided to the LLM includes the dialogue history $C$, along with the four types of commonsense inferences $\mathcal{K}_p$.

\section{Experimentals}
To demonstrate the effectiveness of \textit{Sibyl}, we conduct experiments to answer the following research questions:\\
\noindent \textbf{RQ1}: Does \textit{Sibyl} outperform existing commonsense knowledge when prompting LLMs to generate empathetic responses? \\
\noindent \textbf{RQ2}: Does incorporating \textbf{Dialogue Future} contribute to generating higher-quality empathetic responses? \\
\noindent \textbf{RQ3}: Are all four categories of commonsense knowledge essential for analyzing dialogue context? \\
\noindent \textbf{RQ4}:  Is \textit{Sibyl} effective when applied to different sizes of backbone language models?
\subsection{Datasets}
Our experiments are conducted on the \textsc{EmpatheticDialogues} \citep{EDdata} (ED) and the Emotional Support Conversation \citep{ESCdata} (ESConv). ED is a vast multi-turn dialogue dataset encompassing 25,000 empathetic conversations between a speaker and a listener. 
ESConv comprises approximately 1,053 multi-turn dialogues between a help seeker experiencing emotional distress and a professional supporter. 

\subsection{Implementation Details}

For the implementation of finetuning LLaMA3.1-8B-Instruct and \textit{Flan-t5-xl} models, we utilize the open-source Hugging Face transformers \citep{Transformers}.  We set the learning rate to 3e-5 and training batch size to 16, train up to 5 epochs, and select the best checkpoints based on performance on the validation sets. The whole model is optimized with the Adam \citep{Adam} algorithm. Considering the inference latency of LLaMA models, we employ LoRA-Tuning to finetune only 0.0622\% parameters of the 7B models. Using LoRA not only reduces training costs but also makes our method plug-and-play. The LoRA's rank is set as 8, the $alpha$ is 16, the dropout rate of LoRA is assigned to 0.05, and the target modules are $Q$ and $V$.



\begin{table*}[]
\centering
\scalebox{0.73}{
\begin{tabular}{clccccccc}
\toprule
\textbf{Generation Paradigm} & \textbf{Model} & \textbf{BLEU-1/2/3/4}         & \textbf{Dist-1/2/3}      & \textbf{ROU\_L.} & \textbf{MET.}  & \textbf{Ave.}   & \textbf{Ext.}   & \textbf{CIDEr}  \\ \midrule
& CASE & 16.04/7.5/3.99/2.3 & 0.74/3.03/6.02 & 18.07 & 7.79 & 87.08 & 59.85 & 18.22 \\ 
& LLaMA3.1 & 16.86/5.95/2.7/1.45 & 5.63/\textbf{36.58/72.07} & 15.16 & 7.6 & 87.3 & 48.08 & 13.73  \\
\multirow{2}{*}{Finetuned} & + COMET    & 17.35/6.38/2.89/1.59 & 5.61/35.88/70.83 & 15.31 & 7.74 & 87.29 & 48.45 & 14.47 \\
& + DOCTOR & 17.41/6.32/2.85/1.58 & 5.62/36.18/71.24 & 15.16 & 9.15 & 86.96 & 48.25 & 13.6 \\
& + DIALeCT & 19.57/8.02/4.14/2.42 & 5.53/36.04/70.88 & 17.37 & 8.61 & 87.69 & 49.81 & 22.21
  \\ 
& \textbf{+ \textit{Sibyl}}   & \textbf{21.45/9.35/5.01/2.95*} & \textbf{5.65}/36.11/72.02 & \textbf{19.08*} & \textbf{9.61*} & \textbf{88.36*} & \textbf{50.9} & \textbf{26.93*} \\ \midrule

& GPT-4o & 14.28/5.00/2.28/1.21 & 9.14/39.29/62.85 & 14.01 & 10.66 & 88.90 & 46.66 & 7.90 \\
& + M-Cue CoT & 11.94/3.95/1.67/0.79 & 9.30/39.78/62.47 & 12.64 & 9.29 & 88.31 & 44.66 & 5.72 \\
\multirow{2}{*}{Prompt-based} & + COMET & 14.07/5.06/2.43/1.34 & 9.36/40.13/64.12 & 14.89 & 9.13 & 88.94 & 45.69 & 7.54 \\
& + DOCTOR & 14.55/5.41/2.66/1.49 & 9.68/\textbf{41.92/64.40} & 15.65 & 9.3 & 89.29 & 46.24 & 8.43 \\
& + DIALeCT& 15.36/5.67/2.64/1.39 & 8.98/38.07/60.13 & 16.23 & 9.46 & 89.29 & 47.47 & 10.48 \\
& \textbf{+ \textit{Sibyl}}  &
  \textbf{16.22/6.41/3.21/1.83*} &
  \textbf{9.70}/39.86/62.69 &
  \textbf{17.62*} &
  \textbf{10.09*} &
  \textbf{89.96} &
  \textbf{48.13*} &
  \textbf{14.14*} \\  \bottomrule

\end{tabular}
}
\caption{Automatic Evaluation results on \textsc{EmpatheticDialogues} dataset. 
The best results are highlighted with \textbf{bold}. "*" denotes that the improvement to the best baseline is statistically significant (t-test with $p$-value < 0.01). }
\label{tab: EDauto}
\end{table*}

\begin{table*}[]
\centering
\scalebox{0.73}{
\begin{tabular}{clccccccc}
\toprule
\textbf{Generation Paradigm} & \textbf{Model} & \textbf{BLEU-2/3/4}         & \textbf{Dist-1/2/3}       & \textbf{ROU\_L.}  & \textbf{MET.}  & \textbf{Ave.}   & \textbf{Ext.}   & \textbf{CIDEr}  \\  \midrule
& LLaMA3.1  & 6.75/2.92/1.41   & 6.24/40.34/75.6  & 15.62  & 9.12 & 88.44  & 44.71   & 8.76 \\
& + COMET            & 6.48/2.78/1.35 & 6.22/39.81/75.18 & 15.58          & \textbf{9.04}          & 89.19          & 45             & 9.34 \\
Finetuned & + DOCTOR & 6.58/2.83/1.42 & 6.68/41.32/75.82 & 15.78 & 8.24 & 89.31 & 45.14 & 9.84 \\
& + DIALeCT& 6.78/2.79/1.29 & 6.35/40.46/76.29 & 16.02 & 8.22 & 88.25 & 44.86 & 10.44 \\ 
& \textbf{+ \textit{Sibyl}}  & \textbf{6.97/3.04/1.52*}& \textbf{6.84/41.59/76.41*} & \textbf{16.23} & 8.53 & \textbf{89.55*} & \textbf{45.86} & \textbf{10.92*} \\ \midrule

& GPT-4o & 5.06/2.01/0.93 & 6.43/31.39/56.38 & 14.86          & 8.5           & 90.14 & 41.9           & 4.22 \\
& + M-Cue CoT & 5.03/1.89/0.92 & 6.32/30.97/55.78 & 14.79 & 9.27 & 89.76 & 41.73 & 3.98 \\
\multirow{2}{*}{Prompt-based} & + COMET & 5.06/1.99/0.91 & 5.98/29.56/52.89 & 14.87 & 9.44 & 90.66 & \textbf{42.98} & 4.14 \\
& + DOCTOR &  4.46/1.72/0.79 & 6.36/31.76/56.48 & 13.98 & 8.73 & 90.24 & 40.93 & 3.39\\
& + DIALeCT& 4.95/1.82/0.81 & 6.42/31.14/54.24 & 14.97 & 9.1 & 90.6 & 42.56 & 4.15 \\
& \textbf{+ \textit{Sibyl}}  & \textbf{5.19/2.21/1.10*} & \textbf{6.52/32.09/56.72} & \textbf{15.2*} & \textbf{9.65} & \textbf{90.7*}  & 41.9   & \textbf{4.95} \\ \bottomrule


\end{tabular}
}
\caption{Automatic Evaluation results on ESConv dataset. The best results are highlighted with \textbf{bold}. "*" denotes that the improvement to the best baseline is statistically significant (t-test with $p$-value < 0.01). }
\label{tab: ESCauto}
\end{table*}

\subsection{Baseline Methods}

We compare \textit{Sibyl} with several state-of-the-art methods and commonsense knowledge deduced by other baseline frameworks:\\
\textbf{CASE} \citep{CASE}: A model trained from scratch with vanilla transformers \citep{TRANSFORMER} on ED dataset. This work utilizes a conditional graph to represent all plausible causalities between the user's emotions and experience. \\
\textbf{M-Cue CoT} \citep{Cue-CoT}: A multi-step prompting mechanism to trace the status of users during the conversation, performing complex reasoning and planning before generating the final response.\\
\textbf{LLaMA3.1} \citep{llama3}: To test the performance of vanilla open-source foundation models, we apply LLaMA3.1-8B-Instruct\footnote{The version of LLaMA used in this paper: \url{https://huggingface.co/meta-llama/Meta-Llama-3.1-8B-Instruct}} which only responds based on dialogue context. \\
\textbf{+ COMET} \citep{COMET}: A foundation model enhanced by external knowledge comes from ATOMIC \citep{ATOMIC} which makes inferences based on the last utterance of context.\\
\textbf{+ DOCTOR} \citep{DialogueCoT}: A dialogue Chain-of-Thought commonsense reasoner that integrates implicit information in dialogue into the rationale for generating responses. 
To enhance the quality of the knowledge generated by DOCTOR in our experiments, we further fine-tuned the model using the ED and ESConv training sets, following the guidelines outlined in this paper. \\
\textbf{+ DIALeCT} \citep{CICEROv2}: Trained on a variety of dialogue-related tasks, DIALeCT is a pre-trained transformer for commonsense inference in dialogues which expert in leveraging the structural information from the dialogues.  \\
The detail of the above baseline methods is specified in Appendix \ref{sec:appendix_baselines}. 




\subsection{Automatic Evaluation: RQ1, RQ2}


The generated responses are evaluated using several automatic metrics, namely BLEU \cite{BLEU}, ROUGE-L (\textbf{ROU-L.}) \cite{rouge}, METEOR (\textbf{MET}) \citep{meteor}, Distinct-n (\textbf{Dist}-$n$) \cite{Dist}, and \textbf{CIDEr} \citep{CIDEr}. Additionally, we employ  Average (\textbf{Ave.}) and Extrema (\textbf{Ext.}) Cosine Scores to assess embedding-based semantic similarity. 

Supervised Finetuning (SFT) plays a crucial role in applying LLMs to specific tasks. Our approach significantly outperforms the mentioned baseline methods in generating \textbf{empathetic} responses on both Decoder-Only and Encoder-Decoder models (LLaMA and Flan-t5). As shown in the upper portion of Table \ref{tab: EDauto}, the similarity scores (\textbf{BLEU-n}, \textbf{ROU-L}. and \textbf{MET.}) of responses generated by LLaMA enhanced with \textit{Sibyl} exceed those of all baseline methods by a significant margin, suggesting that the more sensible responses stem from the paradigm’s ability to deduce commonsense knowledge. However, for extrema score (\textbf{Ext.}), \textit{Sibyl} performs slightly worse than the baselines. Equipped with \textit{Sibyl}, LLaMA excels in achieving the highest scores in both average embedding similarity (\textbf{Avg.}) and \textbf{CIDEr}, further proving its effectiveness in empathetic response generation. The performance of the Finetuned model on \textit{Flan-t5-xl}, as depicted in Table \ref{tab: Flan}, additionally shows significant improvement when enhanced by \textit{Sibyl}, especially in the areas of overlap and embedding similarity scores. Impressively, the \textbf{CIDEr} score improvement of our method over the standard model by about 13 points highlights the critical role of anticipating dialogue futures and the distinct effectiveness of our proposed paradigm.

In the context of ESConv, we compared \textit{Sibyl} paradigm to the baseline methods for commonsense knowledge. As shown in Table \ref{tab: ESCauto}, \textit{Sibyl} enhances foundation models' performance in emotional support scenarios. With \textit{Sibyl} integration, LLMs outshine all other categories of commonsense knowledge under diversity metrics (\textbf{Dist-n}), underscoring the critical role of prophetic abilities in response generation. 


\begin{table}[]
\centering
\scalebox{0.72}{
\begin{tabular}{lcccccc}
\toprule
 \textbf{Model} & \textbf{BLEU-3/4}  & \textbf{ROU\_L.} & \textbf{MET.}  & \textbf{Ave.}   & \textbf{CIDEr}  \\ \midrule
 Flan-t5-xl & 5.82/3.78 & 20.73 & 8.92 & 88.35 & 30.44 \\
 + COMET    & 2.49/1.29 & 14.96 & 7.05 & 86.82 & 12.92\\
 + DOCTOR   & 2.58/1.33 & 14.78 & 6.97 & 86.92 & 23.41 \\
 + DIALeCT  & 3.90/2.26 & 17.17 & 8.03 & 87.61 & 13.16\\ 
 \textbf{+ \textit{Sibyl}}  & \textbf{7.71/5.24} & \textbf{23.09} & \textbf{10.39} & \textbf{88.53} & \textbf{43.36} \\ \bottomrule

\end{tabular}
}
\caption{Automatic Evaluation results on \textsc{EmpatheticDialogues} dataset. The foundation model is Flan-t5-xl. The best results are highlighted with \textbf{bold}.}
\label{tab: Flan}
\end{table}

\begin{figure*}
    \centering
    \includegraphics[width=0.99\textwidth]{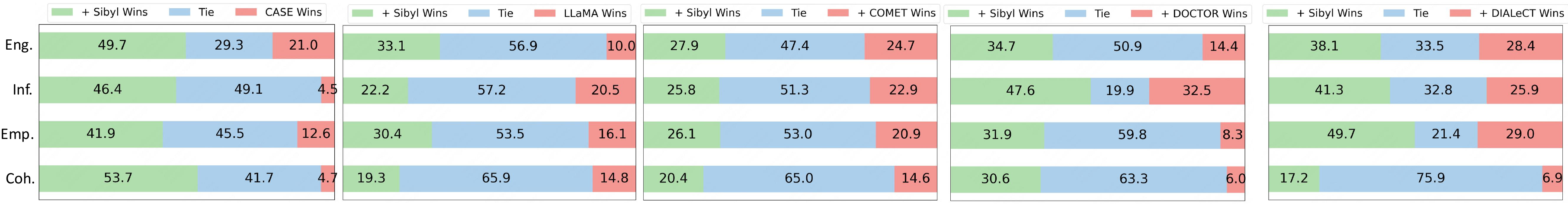}
    \caption{Human A/B test of \textsc{EmpatheticDialogues}($\%$). The results are statistically significant with p-value < 0.05, and \textbf{Kappa} ($\kappa$) falls between 0.4 and 0.6, suggesting moderate agreement among annotators. }
    \label{fig: human_ed}
\end{figure*}

\begin{figure*}
    \centering
    \includegraphics[width=0.99\textwidth]{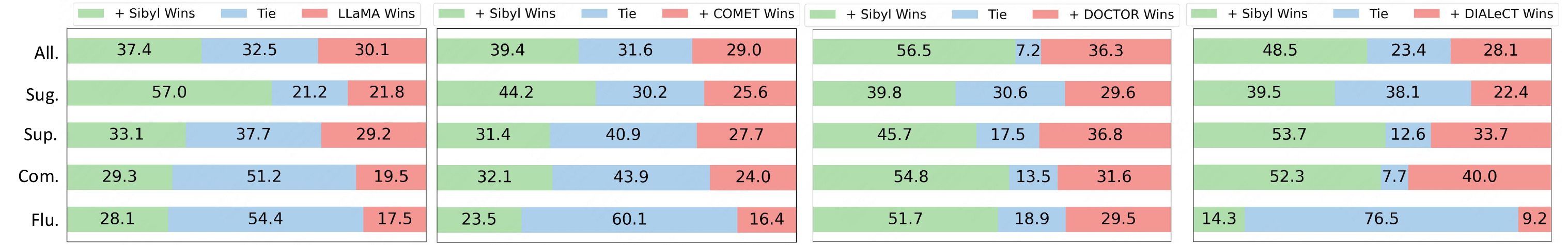}
    \caption{Human A/B test of ESConv ($\%$). The results are statistically significant with p-value < 0.05, and \textbf{Kappa} ($\kappa$) falls between 0.4 and 0.6, suggesting moderate agreement among annotators. }
    \label{fig: human_esc}
\end{figure*}

Given that In-context Learning (ICL) is widely regarded as a key strength of Large Language Models (LLMs), our study explores the influence of different commonsense inferences on LLM response generation without fine-tuning (prompt-based). We focused on conducting experiments using OpenAI's GPT-4o. As outlined in the lower part of Table \ref{tab: EDauto} and Table \ref{tab: ESCauto}, the diversity scores of the content of our methodology generated are competitive with baselines and markedly superior in other metrics for empathetic dialogues. 
In the realm of emotional support, \textit{Sibyl} catalyzes LLMs’ potential to provide empathetic and supportive responses. Through our proposed visionary commonsense inference, LLMs attain scores in Extrema (\textbf{Ext.}) and \textbf{CIDEr} that are on par with the best, while exceeding baseline models in all other diversity-driven and overlapping metrics. Notably, armed with M-Cue CoT GPT-4o performs even worse than prompting without knowledge, which demonstrate the importance of visionary commonsense knowledge in the task of empathetic response generation, as shown in Table \ref{tab: EDauto} and \ref{tab: ESCauto}.

\subsection{Human Evaluation}  \label{human_eval}
For the human evaluation, we focus on four key aspects in the ED dataset: 1) \textit{Coherence} (\textbf{Coh.}): Which model produces responses that are more coherent and relevant to the dialogue context? 2) \textit{Empathy} (\textbf{Emp.}): Which model exhibits more appropriate emotional reactions, such as warmth, compassion, and concern? 3) \textit{Informativeness} (\textbf{Inf.}): Which model provides more contextually relevant information in its responses? 4) \textit{Engagement} (\textbf{Eng.}): Which model’s response is more likely to encourage interlocutors to continue the conversation? Notably, previous works have largely overlooked \textbf{Engagement}, despite its critical role in emulating human-like interactions during real conversations.
 
In the realm of ESConv, we consider five aspects: 1) \textit{Fluency} (\textbf{Flu.}): Evaluating the models based on the fluency of their responses. 2) \textit{Comforting} (\textbf{Com.}): Assessing the models' skill in providing comfort. 3) \textit{Supportive} (\textbf{Sup.}): Determining which model offers more supportive or helpful responses. 4) \textit{Suggestion} (\textbf{Sug.}): which bot gave you more helpful suggestions for your problems? 5) \textit{Overall} (\textbf{All.}): Analyzing which model provides more effective overall emotional support.

\begin{table*}[]
\centering
\scalebox{0.76}{
\begin{tabular}{lccccccc}
\toprule
 \textbf{Model} & \textbf{BLEU-1/2/3/4}         & \textbf{Dist-1/2/3}      & \textbf{ROU\_L.} & \textbf{MET.}  & \textbf{Ave.}   & \textbf{Ext.}   & \textbf{CIDEr}  \\ \midrule
  \textbf{+ \textit{Sibyl}}   & \textbf{21.34/9.25/4.89/2.84} & \textbf{5.61/36.07/71.17} & \textbf{19}   & \textbf{9.54} & \textbf{88.29} & 50.85 & \textbf{26.89} \\ \midrule
\textit{w/o} Cause  & 20.89/9.06/4.78/2.78	& 5.35/34.52/68.48 & 18.69 & 9.38 & 88.01 & \textbf{50.9}  & 25.87\\
 \textit{w/o} Intent & 18.72/7.05/3.35/1.82 & 5.29/33.67/67.44 & 16.18 & 8.17 & 87.34 & 49.12 & 16.46 \\
 \textit{w/o} Subs   & 20.69/8.89/4.66/2.71 & 5.37/34.16/67.91 &	18.23 &	9.2 &	87.83 &	50.45 &	24.39 \\
\textit{w/o} Emo    & 21.18/9.12/4.79/2.74 & 5.41/34.47/68.4 & 18.63 & 9.25 & 87.92 & 50.82 & 25.35  \\  
\bottomrule
\end{tabular}
}
\caption{Ablation study on the ED dataset.}
\label{tab: ED_ablation}
\end{table*}

\begin{table*}[]
\centering
\scalebox{0.76}{
\begin{tabular}{lccccccc}
\toprule
 \textbf{Model} & \textbf{BLEU-2/3/4}         & \textbf{Dist-1/2/3}      & \textbf{ROU\_L.} & \textbf{MET.}  & \textbf{Ave.}   & \textbf{Ext.}   & \textbf{CIDEr}  \\ \midrule
  \textbf{+ \textit{Sibyl}} & \textbf{6.97/3.04/1.52} & \textbf{6.84/41.59/76.41} & \textbf{16.23} & \textbf{8.53} & \textbf{89.55} &\textbf{ 45.86} & 10.92 \\ \midrule
\textit{w/o} Cause  & 6.89/2.99/1.45 & 6.62/41.03/75.69 & 15.99 & 8.45 & 89.01 & 45.59 & \textbf{11.24}\\
 \textit{w/o} Intent & 6.62/2.90/1.42 & 6.50/41.04/75.93 & 15.75 & 8.75 & 89.22 & 45.67 & 10.21 \\
 \textit{w/o} Subs   & 6.74/2.89/1.37 & 6.53/40.99/75.59 & 16.23 & 8.75 & 89.35 & 45.38 & 10.96 \\
\textit{w/o} Emo    & 6.61/2.74/1.25 & 6.62/41.10/75.75 & 15.75 & 8.37 & 89.43 & 45.6 & 10.24  \\  
\bottomrule
\end{tabular}
}
\caption{Ablation study on the ESConv dataset.}
\label{tab: ESC_ablation}
\end{table*}

We randomly select 200 dialogue samples and engage five professional annotators to evaluate the responses generated by finetuned LLaMA models for both the ED and ESConv datasets. The 
Considering the variation between individuals, we conduct human A/B tests to compare our paradigm with other baselines directly. Annotators score the questionnaire of the response pairs to choose one of the responses in random order or select "Tie" when the quality of those provided sentences is difficult to distinguish. Figure \ref{fig: human_ed} demonstrates \textit{Sibyl}'s advantage over CASE across all metrics. Compared to commonsense inference obtained from COMET, DOCTOR, and DIALeCT, our paradigm exhibits considerable progress, highlighting our approach's effectiveness in incorporating commonsense knowledge. These comparisons emphasize our paradigm's superior performance compared to the three baseline commonsense knowledge. Similarly, results from Figure \ref{fig: human_esc} strongly highlight the effectiveness of \textit{Sibyl} within emotional support scenarios. The considerable lead in the overall score over the baselines indicates a more substantial influence, demonstrating the greater supportiveness of the knowledge, acting as cues that guide LLMs to be more helpful. 

\subsection{Ablation Study: RQ3}
To assess the influence of different categories of commonsense knowledge on response generation and address \textbf{Q3}, we systematically remove each of these four categories of commonsense knowledge to facilitate a performance comparison on the ED and ESConv dataset with \textit{Sibyl}, as illustrated in Table \ref{tab: ED_ablation} and \ref{tab: ESC_ablation}. Excluding any of the four commonsense knowledge categories leads to a reduction in the quality of the generated response. Although some variants perform better than the complete method in particular metrics, the overall performance shows a notable decrease. The causality of the conversation holds less significance in the generation of empathetic responses, whereas emotional cues provide greater insight into future information for understanding the user's situation.
Furthermore, the conspicuous disparity between the variant (\textit{w/o intent}) and our proposed complete method highlights the importance of predicting the potential intent of future responses, aligning with earlier studies \citep{EMPAINTENT, SEEK}.

\subsection{LLMs-based Evaluation}

We apply G-Eval \citep{G-EVAL} to assess the Naturalness (\textbf{Nat.}) and Coherence (\textbf{Coh.}) of responses from baseline approaches that utilize commonsense knowledge in diverse ways. For task-specific requirements, we compare Empathy (\textbf{Emp.}) in the context of \textsc{EmpatheticDialogues} and Supportiveness (\textbf{Sup.}) for ESConv. 

We randomly selected 200 data from both ED and ESConv datasets to perform G-Eval evaluation. Calculating the average weighted score of sampled data, the comparison result is shown in Table \ref{tab: G_eval_SFT} and Table \ref{tab: G_eval_ICL}, \textit{Sibyl} outperforms all strong baseline of commonsense inference in all aspects. The prompt template is specified in Figure \ref{Geval_template}.
More details are provided in Appendix  \ref{sec:appendix_GEVal}.

\subsection{\textit{Sibyl} on Small Language Models: RQ4}
To further validate the effectiveness of \textit{Sibyl}, we augmented the Normal Transformer with commonsense knowledge from \textit{Sibyl} for the multi-turn ED dataset. By merely appending four categories of commonsense knowledge to the dialogue context, without any task instructions or background information, the results, shown in Table \ref{tab: ED_backbone}, demonstrate that Normal TRS + \textit{Sibyl} surpasses both CASE and EmpSOA in automated metrics. 
Similarly, we incorporated \textit{Sibyl} into not only the Normal Transformer but also Blenderbot-small-90M \citep{Blenderbot} and Bart-base \citep{BART}, which are widely used backbones on the ESConv dataset. The comparison results are shown in Table \ref{tab: ESC_backbone}. As MISC \citep{MISC} and MultiESC \citep{MultiESC} utilize commonsense knowledge generated by COMET and emotion cause as additional information and are equipped with sophisticated modules to encode external commonsense knowledge and semantic causal information, \textit{Sibyl} surpass these strong backbones by simply appending commonsense inference to the dialogue context on the above-mentioned backbones. 

\subsection{Case Study}
To better evaluate the performance of response generation, we selected an example generated by our proposed paradigm and baselines for comparison. The example in Table \ref{tab: casestudy} demonstrates that baseline models employing COMET and DIALeCT to derive commonsense knowledge struggled to identify the future direction of the dialogue. Although DOCTOR partially recognized the potential information about the future to some extent, these three kinds of inferences still led to responses that were deficient in coherence and empathy. In contrast, \textit{Sibyl} concentrates on crucial information, such as the possibility of the speaker having regular interactions with children. The visionary red-highlighted words accurately identify this detailed information, leading to a more sensible and suggestive response.

\section{Conclusion}
Even when enhanced with commonsense knowledge, LLMs still struggle with providing sensible and empathetic responses when providing support in multi-turn conversations. This paper posits that the underlying issue stems from the one-to-many nature of dialogue generation and commonsense inference. 
We introduce a simple but effective paradigm named \textit{Sibyl}, bridging the gap between context and intended response, and aiding different scales of foundation models to envision dialogue future. 
Through rigorous evaluation, \textit{Sibyl} has demonstrated its superiority as a model-agnostic approach, evidenced by notable improvements in automated metrics and assessments conducted by human evaluators and advanced LLMs.

\section*{Limitations}
In this paper, we introduce a new paradigm for acquiring visionary commonsense knowledge, called Sibyl. However, evaluating empathetic dialogue systems remains challenging.
As highlighted by \citet{HowNot}, the scores from automatic evaluation metrics often do not align fully with human assessments in dialogue generation tasks. Leveraging large language models (LLMs) as expert assessors helps mitigate the issue of lacking labor-free, task-specific evaluation metrics.
Despite this, evaluating empathy and supportiveness in generated content automatically and convincingly remains problematic. To address these challenges, we employ a comprehensive evaluation strategy that incorporates all three methods: automatic evaluation, LLM-based assessment, and human evaluation. This approach ensures a thorough assessment of the responses and validates the effectiveness of our proposed method.



\section*{Ethics Statement}
The datasets \citep{EDdata, ESCdata} utilized in our study are widely recognized and sourced exclusively from open-source repositories.  The conversations of the ED dataset are around given emotions and carried out by employed crowd-sourced workers, with no personal privacy issues involved. 

For our human evaluation, all participants we've recruited are experienced in assessing the quality of responses generated by empathetic dialogue systems. They are knowledgeable about the concept of empathy as defined within the field of psychological academic research. Moreover, participants were provided with fair and appropriate compensation for their involvement. 

\section*{Acknowledgments}
This work was supported by the National Natural Science Foundation of China (No. 62472419). 

\quad \\
The authors extend their gratitude to Naibin Gu from the Institute of Information Engineering, Xiaolei Huang from the University of Memphis, and Yuchen Lin from The University of Hong Kong for their invaluable assistance. We also thank the anonymous reviewers for their insightful feedback, which greatly contributed to improving our paper.

\bibliography{anthology}

\begin{thebibliography}{55}
\providecommand{\natexlab}[1]{#1}

\bibitem[{Bosselut et~al.(2019)Bosselut, Rashkin, Sap, Malaviya, Celikyilmaz, and Choi}]{COMET}
Antoine Bosselut, Hannah Rashkin, Maarten Sap, Chaitanya Malaviya, Asli Celikyilmaz, and Yejin Choi. 2019.
\newblock \href {https://doi.org/10.18653/v1/p19-1470} {{COMET:} commonsense transformers for automatic knowledge graph construction}.
\newblock In \emph{Proceedings of the 57th Conference of the Association for Computational Linguistics, {ACL} 2019, Florence, Italy, July 28- August 2, 2019, Volume 1: Long Papers}, pages 4762--4779. Association for Computational Linguistics.

\bibitem[{Chae et~al.(2023)Chae, Song, iunn Ong, Kwon, Kim, Yu, Lee, Kang, and Yeo}]{DialogueCoT}
Hyungjoo Chae, Yongho Song, Kai~Tzu iunn Ong, Taeyoon Kwon, Minjin Kim, Youngjae Yu, Dongha Lee, Dongyeop Kang, and Jinyoung Yeo. 2023.
\newblock \href {https://arxiv.org/abs/2310.09343} {Dialogue chain-of-thought distillation for commonsense-aware conversational agents}.
\newblock \emph{Preprint}, arXiv:2310.09343.

\bibitem[{Chen et~al.(2022)Chen, Li, and Yang}]{EMPAINTENT}
Mao~Yan Chen, Siheng Li, and Yujiu Yang. 2022.
\newblock \href {https://doi.org/10.18653/v1/2022.naacl-main.78} {Emphi: Generating empathetic responses with human-like intents}.
\newblock In \emph{Proceedings of the 2022 Conference of the North American Chapter of the Association for Computational Linguistics: Human Language Technologies, {NAACL} 2022, Seattle, WA, United States, July 10-15, 2022}, pages 1063--1074. Association for Computational Linguistics.

\bibitem[{Cheng et~al.(2022)Cheng, Liu, Li, Wang, Zhao, Liu, Liang, and Zheng}]{MultiESC}
Yi~Cheng, Wenge Liu, Wenjie Li, Jiashuo Wang, Ruihui Zhao, Bang Liu, Xiaodan Liang, and Yefeng Zheng. 2022.
\newblock \href {https://doi.org/10.18653/v1/2022.emnlp-main.195} {Improving multi-turn emotional support dialogue generation with lookahead strategy planning}.
\newblock In \emph{Proceedings of the 2022 Conference on Empirical Methods in Natural Language Processing, {EMNLP} 2022, Abu Dhabi, United Arab Emirates, December 7-11, 2022}, pages 3014--3026. Association for Computational Linguistics.

\bibitem[{Chiang and yi~Lee(2023)}]{closergeval}
Cheng-Han Chiang and Hung yi~Lee. 2023.
\newblock \href {https://arxiv.org/abs/2310.05657} {A closer look into automatic evaluation using large language models}.
\newblock \emph{Preprint}, arXiv:2310.05657.

\bibitem[{Chlopan et~al.(1985)Chlopan, McCain, Carbonell, and Hagen}]{chlopan1985empathy}
Bruce~E Chlopan, Marianne~L McCain, Joyce~L Carbonell, and Richard~L Hagen. 1985.
\newblock Empathy: Review of available measures.
\newblock \emph{Journal of personality and social psychology}, 48(3):635.

\bibitem[{Chung et~al.(2022)Chung, Hou, Longpre, Zoph, Tay, Fedus, Li, Wang, Dehghani, Brahma, Webson, Gu, Dai, Suzgun, Chen, Chowdhery, Narang, Mishra, Yu, Zhao, Huang, Dai, Yu, Petrov, Chi, Dean, Devlin, Roberts, Zhou, Le, and Wei}]{flant5}
Hyung~Won Chung, Le~Hou, Shayne Longpre, Barret Zoph, Yi~Tay, William Fedus, Eric Li, Xuezhi Wang, Mostafa Dehghani, Siddhartha Brahma, Albert Webson, Shixiang~Shane Gu, Zhuyun Dai, Mirac Suzgun, Xinyun Chen, Aakanksha Chowdhery, Sharan Narang, Gaurav Mishra, Adams Yu, Vincent~Y. Zhao, Yanping Huang, Andrew~M. Dai, Hongkun Yu, Slav Petrov, Ed~H. Chi, Jeff Dean, Jacob Devlin, Adam Roberts, Denny Zhou, Quoc~V. Le, and Jason Wei. 2022.
\newblock \href {https://doi.org/10.48550/ARXIV.2210.11416} {Scaling instruction-finetuned language models}.
\newblock \emph{CoRR}, abs/2210.11416.

\bibitem[{Davis(1983)}]{davis1983measuring}
Mark~H Davis. 1983.
\newblock Measuring individual differences in empathy: Evidence for a multidimensional approach.
\newblock \emph{Journal of personality and social psychology}, 44(1):113.

\bibitem[{Deng et~al.(2023)Deng, Zhang, Yuan, and Lam}]{KEMI}
Yang Deng, Wenxuan Zhang, Yifei Yuan, and Wai Lam. 2023.
\newblock \href {https://doi.org/10.18653/V1/2023.ACL-LONG.225} {Knowledge-enhanced mixed-initiative dialogue system for emotional support conversations}.
\newblock In \emph{Proceedings of the 61st Annual Meeting of the Association for Computational Linguistics (Volume 1: Long Papers), {ACL} 2023, Toronto, Canada, July 9-14, 2023}, pages 4079--4095. Association for Computational Linguistics.

\bibitem[{Dubey et~al.(2024)Dubey, Jauhri, Pandey, Kadian, Al-Dahle, Letman, Mathur, Schelten, Yang, Fan, Goyal, Hartshorn, Yang, Mitra, Sravankumar, Korenev, Hinsvark, Rao, Zhang, Rodriguez, Gregerson, Spataru, Roziere, Biron, Tang, Chern, Caucheteux, Nayak, Bi, Marra, McConnell, Keller, Touret, Wu, Wong, Ferrer, Nikolaidis, Allonsius, Song, Pintz, Livshits, Esiobu, Choudhary, Mahajan, Garcia-Olano, Perino, Hupkes, Lakomkin, AlBadawy, Lobanova, Dinan, Smith, Radenovic, Zhang, Synnaeve, Lee, Anderson, Nail, Mialon, Pang, Cucurell, Nguyen, Korevaar, Xu, Touvron, Zarov, Ibarra, Kloumann, Misra, Evtimov, Copet, Lee, Geffert, Vranes, Park, Mahadeokar, Shah, van~der Linde, Billock, Hong, Lee, Fu, Chi, Huang, Liu, Wang, Yu, Bitton, Spisak, Park, Rocca, Johnstun, Saxe, Jia, Alwala, Upasani, Plawiak, Li, Heafield, Stone, El-Arini, Iyer, Malik, Chiu, Bhalla, Rantala-Yeary, van~der Maaten, Chen, Tan, Jenkins, Martin, Madaan, Malo, Blecher, Landzaat, de~Oliveira, Muzzi, Pasupuleti, Singh, Paluri, Kardas, Oldham, Rita,
  Pavlova, Kambadur, Lewis, Si, Singh, Hassan, Goyal, Torabi, Bashlykov, Bogoychev, Chatterji, Duchenne, Çelebi, Alrassy, Zhang, Li, Vasic, Weng, Bhargava, Dubal, Krishnan, Koura, Xu, He, Dong, Srinivasan, Ganapathy, Calderer, Cabral, Stojnic, Raileanu, Girdhar, Patel, Sauvestre, Polidoro, Sumbaly, Taylor, Silva, Hou, Wang, Hosseini, Chennabasappa, Singh, Bell, Kim, Edunov, Nie, Narang, Raparthy, Shen, Wan, Bhosale, Zhang, Vandenhende, Batra, Whitman, Sootla, Collot, Gururangan, Borodinsky, Herman, Fowler, Sheasha, Georgiou, Scialom, Speckbacher, Mihaylov, Xiao, Karn, Goswami, Gupta, Ramanathan, Kerkez, Gonguet, Do, Vogeti, Petrovic, Chu, Xiong, Fu, Meers, Martinet, Wang, Tan, Xie, Jia, Wang, Goldschlag, Gaur, Babaei, Wen, Song, Zhang, Li, Mao, Coudert, Yan, Chen, Papakipos, Singh, Grattafiori, Jain, Kelsey, Shajnfeld, Gangidi, Victoria, Goldstand, Menon, Sharma, Boesenberg, Vaughan, Baevski, Feinstein, Kallet, Sangani, Yunus, Lupu, Alvarado, Caples, Gu, Ho, Poulton, Ryan, Ramchandani, Franco, Saraf,
  Chowdhury, Gabriel, Bharambe, Eisenman, Yazdan, James, Maurer, Leonhardi, Huang, Loyd, Paola, Paranjape, Liu, Wu, Ni, Hancock, Wasti, Spence, Stojkovic, Gamido, Montalvo, Parker, Burton, Mejia, Wang, Kim, Zhou, Hu, Chu, Cai, Tindal, Feichtenhofer, Civin, Beaty, Kreymer, Li, Wyatt, Adkins, Xu, Testuggine, David, Parikh, Liskovich, Foss, Wang, Le, Holland, Dowling, Jamil, Montgomery, Presani, Hahn, Wood, Brinkman, Arcaute, Dunbar, Smothers, Sun, Kreuk, Tian, Ozgenel, Caggioni, Guzmán, Kanayet, Seide, Florez, Schwarz, Badeer, Swee, Halpern, Thattai, Herman, Sizov, Guangyi, Zhang, Lakshminarayanan, Shojanazeri, Zou, Wang, Zha, Habeeb, Rudolph, Suk, Aspegren, Goldman, Damlaj, Molybog, Tufanov, Veliche, Gat, Weissman, Geboski, Kohli, Asher, Gaya, Marcus, Tang, Chan, Zhen, Reizenstein, Teboul, Zhong, Jin, Yang, Cummings, Carvill, Shepard, McPhie, Torres, Ginsburg, Wang, Wu, U, Saxena, Prasad, Khandelwal, Zand, Matosich, Veeraraghavan, Michelena, Li, Huang, Chawla, Lakhotia, Huang, Chen, Garg, A, Silva, Bell,
  Zhang, Guo, Yu, Moshkovich, Wehrstedt, Khabsa, Avalani, Bhatt, Tsimpoukelli, Mankus, Hasson, Lennie, Reso, Groshev, Naumov, Lathi, Keneally, Seltzer, Valko, Restrepo, Patel, Vyatskov, Samvelyan, Clark, Macey, Wang, Hermoso, Metanat, Rastegari, Bansal, Santhanam, Parks, White, Bawa, Singhal, Egebo, Usunier, Laptev, Dong, Zhang, Cheng, Chernoguz, Hart, Salpekar, Kalinli, Kent, Parekh, Saab, Balaji, Rittner, Bontrager, Roux, Dollar, Zvyagina, Ratanchandani, Yuvraj, Liang, Alao, Rodriguez, Ayub, Murthy, Nayani, Mitra, Li, Hogan, Battey, Wang, Maheswari, Howes, Rinott, Bondu, Datta, Chugh, Hunt, Dhillon, Sidorov, Pan, Verma, Yamamoto, Ramaswamy, Lindsay, Lindsay, Feng, Lin, Zha, Shankar, Zhang, Zhang, Wang, Agarwal, Sajuyigbe, Chintala, Max, Chen, Kehoe, Satterfield, Govindaprasad, Gupta, Cho, Virk, Subramanian, Choudhury, Goldman, Remez, Glaser, Best, Kohler, Robinson, Li, Zhang, Matthews, Chou, Shaked, Vontimitta, Ajayi, Montanez, Mohan, Kumar, Mangla, Albiero, Ionescu, Poenaru, Mihailescu, Ivanov, Li, Wang,
  Jiang, Bouaziz, Constable, Tang, Wang, Wu, Wang, Xia, Wu, Gao, Chen, Hu, Jia, Qi, Li, Zhang, Zhang, Adi, Nam, Yu, Wang, Hao, Qian, He, Rait, DeVito, Rosnbrick, Wen, Yang, and Zhao}]{llama3}
Abhimanyu Dubey, Abhinav Jauhri, Abhinav Pandey, Abhishek Kadian, Ahmad Al-Dahle, Aiesha Letman, Akhil Mathur, Alan Schelten, Amy Yang, Angela Fan, Anirudh Goyal, Anthony Hartshorn, Aobo Yang, Archi Mitra, Archie Sravankumar, Artem Korenev, Arthur Hinsvark, Arun Rao, Aston Zhang, Aurelien Rodriguez, Austen Gregerson, Ava Spataru, Baptiste Roziere, Bethany Biron, Binh Tang, Bobbie Chern, Charlotte Caucheteux, Chaya Nayak, Chloe Bi, Chris Marra, Chris McConnell, Christian Keller, Christophe Touret, Chunyang Wu, Corinne Wong, Cristian~Canton Ferrer, Cyrus Nikolaidis, Damien Allonsius, Daniel Song, Danielle Pintz, Danny Livshits, David Esiobu, Dhruv Choudhary, Dhruv Mahajan, Diego Garcia-Olano, Diego Perino, Dieuwke Hupkes, Egor Lakomkin, Ehab AlBadawy, Elina Lobanova, Emily Dinan, Eric~Michael Smith, Filip Radenovic, Frank Zhang, Gabriel Synnaeve, Gabrielle Lee, Georgia~Lewis Anderson, Graeme Nail, Gregoire Mialon, Guan Pang, Guillem Cucurell, Hailey Nguyen, Hannah Korevaar, Hu~Xu, Hugo Touvron, Iliyan Zarov,
  Imanol~Arrieta Ibarra, Isabel Kloumann, Ishan Misra, Ivan Evtimov, Jade Copet, Jaewon Lee, Jan Geffert, Jana Vranes, Jason Park, Jay Mahadeokar, Jeet Shah, Jelmer van~der Linde, Jennifer Billock, Jenny Hong, Jenya Lee, Jeremy Fu, Jianfeng Chi, Jianyu Huang, Jiawen Liu, Jie Wang, Jiecao Yu, Joanna Bitton, Joe Spisak, Jongsoo Park, Joseph Rocca, Joshua Johnstun, Joshua Saxe, Junteng Jia, Kalyan~Vasuden Alwala, Kartikeya Upasani, Kate Plawiak, Ke~Li, Kenneth Heafield, Kevin Stone, Khalid El-Arini, Krithika Iyer, Kshitiz Malik, Kuenley Chiu, Kunal Bhalla, Lauren Rantala-Yeary, Laurens van~der Maaten, Lawrence Chen, Liang Tan, Liz Jenkins, Louis Martin, Lovish Madaan, Lubo Malo, Lukas Blecher, Lukas Landzaat, Luke de~Oliveira, Madeline Muzzi, Mahesh Pasupuleti, Mannat Singh, Manohar Paluri, Marcin Kardas, Mathew Oldham, Mathieu Rita, Maya Pavlova, Melanie Kambadur, Mike Lewis, Min Si, Mitesh~Kumar Singh, Mona Hassan, Naman Goyal, Narjes Torabi, Nikolay Bashlykov, Nikolay Bogoychev, Niladri Chatterji, Olivier
  Duchenne, Onur Çelebi, Patrick Alrassy, Pengchuan Zhang, Pengwei Li, Petar Vasic, Peter Weng, Prajjwal Bhargava, Pratik Dubal, Praveen Krishnan, Punit~Singh Koura, Puxin Xu, Qing He, Qingxiao Dong, Ragavan Srinivasan, Raj Ganapathy, Ramon Calderer, Ricardo~Silveira Cabral, Robert Stojnic, Roberta Raileanu, Rohit Girdhar, Rohit Patel, Romain Sauvestre, Ronnie Polidoro, Roshan Sumbaly, Ross Taylor, Ruan Silva, Rui Hou, Rui Wang, Saghar Hosseini, Sahana Chennabasappa, Sanjay Singh, Sean Bell, Seohyun~Sonia Kim, Sergey Edunov, Shaoliang Nie, Sharan Narang, Sharath Raparthy, Sheng Shen, Shengye Wan, Shruti Bhosale, Shun Zhang, Simon Vandenhende, Soumya Batra, Spencer Whitman, Sten Sootla, Stephane Collot, Suchin Gururangan, Sydney Borodinsky, Tamar Herman, Tara Fowler, Tarek Sheasha, Thomas Georgiou, Thomas Scialom, Tobias Speckbacher, Todor Mihaylov, Tong Xiao, Ujjwal Karn, Vedanuj Goswami, Vibhor Gupta, Vignesh Ramanathan, Viktor Kerkez, Vincent Gonguet, Virginie Do, Vish Vogeti, Vladan Petrovic, Weiwei Chu,
  Wenhan Xiong, Wenyin Fu, Whitney Meers, Xavier Martinet, Xiaodong Wang, Xiaoqing~Ellen Tan, Xinfeng Xie, Xuchao Jia, Xuewei Wang, Yaelle Goldschlag, Yashesh Gaur, Yasmine Babaei, Yi~Wen, Yiwen Song, Yuchen Zhang, Yue Li, Yuning Mao, Zacharie~Delpierre Coudert, Zheng Yan, Zhengxing Chen, Zoe Papakipos, Aaditya Singh, Aaron Grattafiori, Abha Jain, Adam Kelsey, Adam Shajnfeld, Adithya Gangidi, Adolfo Victoria, Ahuva Goldstand, Ajay Menon, Ajay Sharma, Alex Boesenberg, Alex Vaughan, Alexei Baevski, Allie Feinstein, Amanda Kallet, Amit Sangani, Anam Yunus, Andrei Lupu, Andres Alvarado, Andrew Caples, Andrew Gu, Andrew Ho, Andrew Poulton, Andrew Ryan, Ankit Ramchandani, Annie Franco, Aparajita Saraf, Arkabandhu Chowdhury, Ashley Gabriel, Ashwin Bharambe, Assaf Eisenman, Azadeh Yazdan, Beau James, Ben Maurer, Benjamin Leonhardi, Bernie Huang, Beth Loyd, Beto~De Paola, Bhargavi Paranjape, Bing Liu, Bo~Wu, Boyu Ni, Braden Hancock, Bram Wasti, Brandon Spence, Brani Stojkovic, Brian Gamido, Britt Montalvo, Carl
  Parker, Carly Burton, Catalina Mejia, Changhan Wang, Changkyu Kim, Chao Zhou, Chester Hu, Ching-Hsiang Chu, Chris Cai, Chris Tindal, Christoph Feichtenhofer, Damon Civin, Dana Beaty, Daniel Kreymer, Daniel Li, Danny Wyatt, David Adkins, David Xu, Davide Testuggine, Delia David, Devi Parikh, Diana Liskovich, Didem Foss, Dingkang Wang, Duc Le, Dustin Holland, Edward Dowling, Eissa Jamil, Elaine Montgomery, Eleonora Presani, Emily Hahn, Emily Wood, Erik Brinkman, Esteban Arcaute, Evan Dunbar, Evan Smothers, Fei Sun, Felix Kreuk, Feng Tian, Firat Ozgenel, Francesco Caggioni, Francisco Guzmán, Frank Kanayet, Frank Seide, Gabriela~Medina Florez, Gabriella Schwarz, Gada Badeer, Georgia Swee, Gil Halpern, Govind Thattai, Grant Herman, Grigory Sizov, Guangyi, Zhang, Guna Lakshminarayanan, Hamid Shojanazeri, Han Zou, Hannah Wang, Hanwen Zha, Haroun Habeeb, Harrison Rudolph, Helen Suk, Henry Aspegren, Hunter Goldman, Ibrahim Damlaj, Igor Molybog, Igor Tufanov, Irina-Elena Veliche, Itai Gat, Jake Weissman, James
  Geboski, James Kohli, Japhet Asher, Jean-Baptiste Gaya, Jeff Marcus, Jeff Tang, Jennifer Chan, Jenny Zhen, Jeremy Reizenstein, Jeremy Teboul, Jessica Zhong, Jian Jin, Jingyi Yang, Joe Cummings, Jon Carvill, Jon Shepard, Jonathan McPhie, Jonathan Torres, Josh Ginsburg, Junjie Wang, Kai Wu, Kam~Hou U, Karan Saxena, Karthik Prasad, Kartikay Khandelwal, Katayoun Zand, Kathy Matosich, Kaushik Veeraraghavan, Kelly Michelena, Keqian Li, Kun Huang, Kunal Chawla, Kushal Lakhotia, Kyle Huang, Lailin Chen, Lakshya Garg, Lavender A, Leandro Silva, Lee Bell, Lei Zhang, Liangpeng Guo, Licheng Yu, Liron Moshkovich, Luca Wehrstedt, Madian Khabsa, Manav Avalani, Manish Bhatt, Maria Tsimpoukelli, Martynas Mankus, Matan Hasson, Matthew Lennie, Matthias Reso, Maxim Groshev, Maxim Naumov, Maya Lathi, Meghan Keneally, Michael~L. Seltzer, Michal Valko, Michelle Restrepo, Mihir Patel, Mik Vyatskov, Mikayel Samvelyan, Mike Clark, Mike Macey, Mike Wang, Miquel~Jubert Hermoso, Mo~Metanat, Mohammad Rastegari, Munish Bansal, Nandhini
  Santhanam, Natascha Parks, Natasha White, Navyata Bawa, Nayan Singhal, Nick Egebo, Nicolas Usunier, Nikolay~Pavlovich Laptev, Ning Dong, Ning Zhang, Norman Cheng, Oleg Chernoguz, Olivia Hart, Omkar Salpekar, Ozlem Kalinli, Parkin Kent, Parth Parekh, Paul Saab, Pavan Balaji, Pedro Rittner, Philip Bontrager, Pierre Roux, Piotr Dollar, Polina Zvyagina, Prashant Ratanchandani, Pritish Yuvraj, Qian Liang, Rachad Alao, Rachel Rodriguez, Rafi Ayub, Raghotham Murthy, Raghu Nayani, Rahul Mitra, Raymond Li, Rebekkah Hogan, Robin Battey, Rocky Wang, Rohan Maheswari, Russ Howes, Ruty Rinott, Sai~Jayesh Bondu, Samyak Datta, Sara Chugh, Sara Hunt, Sargun Dhillon, Sasha Sidorov, Satadru Pan, Saurabh Verma, Seiji Yamamoto, Sharadh Ramaswamy, Shaun Lindsay, Shaun Lindsay, Sheng Feng, Shenghao Lin, Shengxin~Cindy Zha, Shiva Shankar, Shuqiang Zhang, Shuqiang Zhang, Sinong Wang, Sneha Agarwal, Soji Sajuyigbe, Soumith Chintala, Stephanie Max, Stephen Chen, Steve Kehoe, Steve Satterfield, Sudarshan Govindaprasad, Sumit Gupta,
  Sungmin Cho, Sunny Virk, Suraj Subramanian, Sy~Choudhury, Sydney Goldman, Tal Remez, Tamar Glaser, Tamara Best, Thilo Kohler, Thomas Robinson, Tianhe Li, Tianjun Zhang, Tim Matthews, Timothy Chou, Tzook Shaked, Varun Vontimitta, Victoria Ajayi, Victoria Montanez, Vijai Mohan, Vinay~Satish Kumar, Vishal Mangla, Vítor Albiero, Vlad Ionescu, Vlad Poenaru, Vlad~Tiberiu Mihailescu, Vladimir Ivanov, Wei Li, Wenchen Wang, Wenwen Jiang, Wes Bouaziz, Will Constable, Xiaocheng Tang, Xiaofang Wang, Xiaojian Wu, Xiaolan Wang, Xide Xia, Xilun Wu, Xinbo Gao, Yanjun Chen, Ye~Hu, Ye~Jia, Ye~Qi, Yenda Li, Yilin Zhang, Ying Zhang, Yossi Adi, Youngjin Nam, Yu, Wang, Yuchen Hao, Yundi Qian, Yuzi He, Zach Rait, Zachary DeVito, Zef Rosnbrick, Zhaoduo Wen, Zhenyu Yang, and Zhiwei Zhao. 2024.
\newblock \href {https://arxiv.org/abs/2407.21783} {The llama 3 herd of models}.
\newblock \emph{Preprint}, arXiv:2407.21783.

\bibitem[{Fu et~al.(2023)Fu, Ng, Jiang, and Liu}]{GPTscore}
Jinlan Fu, See{-}Kiong Ng, Zhengbao Jiang, and Pengfei Liu. 2023.
\newblock \href {https://doi.org/10.48550/arXiv.2302.04166} {Gptscore: Evaluate as you desire}.
\newblock \emph{CoRR}, abs/2302.04166.

\bibitem[{Ghosal et~al.(2022)Ghosal, Shen, Majumder, Mihalcea, and Poria}]{CICEROv1}
Deepanway Ghosal, Siqi Shen, Navonil Majumder, Rada Mihalcea, and Soujanya Poria. 2022.
\newblock \href {https://doi.org/10.18653/v1/2022.acl-long.344} {{CICERO:} {A} dataset for contextualized commonsense inference in dialogues}.
\newblock In \emph{Proceedings of the 60th Annual Meeting of the Association for Computational Linguistics (Volume 1: Long Papers), {ACL} 2022, Dublin, Ireland, May 22-27, 2022}, pages 5010--5028. Association for Computational Linguistics.

\bibitem[{Han et~al.(2023)Han, Du, Zhang, Lian, Li, Gao, and Wang}]{DiaCoT_and_PPO}
Chengcheng Han, Xiaowei Du, Che Zhang, Yixin Lian, Xiang Li, Ming Gao, and Baoyuan Wang. 2023.
\newblock \href {https://doi.org/10.18653/V1/2023.EMNLP-MAIN.501} {Dialcot meets {PPO:} decomposing and exploring reasoning paths in smaller language models}.
\newblock In \emph{Proceedings of the 2023 Conference on Empirical Methods in Natural Language Processing, {EMNLP} 2023, Singapore, December 6-10, 2023}, pages 8055--8068. Association for Computational Linguistics.

\bibitem[{Hwang et~al.(2021)Hwang, Bhagavatula, Bras, Da, Sakaguchi, Bosselut, and Choi}]{ATOMIC}
Jena~D. Hwang, Chandra Bhagavatula, Ronan~Le Bras, Jeff Da, Keisuke Sakaguchi, Antoine Bosselut, and Yejin Choi. 2021.
\newblock \href {https://doi.org/10.1609/aaai.v35i7.16792} {(comet-) atomic 2020: On symbolic and neural commonsense knowledge graphs}.
\newblock In \emph{Thirty-Fifth {AAAI} Conference on Artificial Intelligence, {AAAI} 2021, Thirty-Third Conference on Innovative Applications of Artificial Intelligence, {IAAI} 2021, The Eleventh Symposium on Educational Advances in Artificial Intelligence, {EAAI} 2021, Virtual Event, February 2-9, 2021}, pages 6384--6392. {AAAI} Press.

\bibitem[{Keskin(2014)}]{keskin2014isn}
Sevgi~Co{\c{s}}kun Keskin. 2014.
\newblock From what isn’t empathy to empathic learning process.
\newblock \emph{Procedia-Social and Behavioral Sciences}, 116:4932--4938.

\bibitem[{Kingma and Ba(2015)}]{Adam}
Diederik~P. Kingma and Jimmy Ba. 2015.
\newblock \href {http://arxiv.org/abs/1412.6980} {Adam: {A} method for stochastic optimization}.
\newblock In \emph{3rd International Conference on Learning Representations, {ICLR} 2015, San Diego, CA, USA, May 7-9, 2015, Conference Track Proceedings}.

\bibitem[{Lavie and Agarwal(2007)}]{meteor}
Alon Lavie and Abhaya Agarwal. 2007.
\newblock \href {https://aclanthology.org/W07-0734/} {{METEOR:} an automatic metric for {MT} evaluation with high levels of correlation with human judgments}.
\newblock In \emph{Proceedings of the Second Workshop on Statistical Machine Translation, WMT@ACL 2007, Prague, Czech Republic, June 23, 2007}, pages 228--231. Association for Computational Linguistics.

\bibitem[{Lewis et~al.(2020)Lewis, Liu, Goyal, Ghazvininejad, Mohamed, Levy, Stoyanov, and Zettlemoyer}]{BART}
Mike Lewis, Yinhan Liu, Naman Goyal, Marjan Ghazvininejad, Abdelrahman Mohamed, Omer Levy, Veselin Stoyanov, and Luke Zettlemoyer. 2020.
\newblock \href {https://doi.org/10.18653/V1/2020.ACL-MAIN.703} {{BART:} denoising sequence-to-sequence pre-training for natural language generation, translation, and comprehension}.
\newblock In \emph{Proceedings of the 58th Annual Meeting of the Association for Computational Linguistics, {ACL} 2020, Online, July 5-10, 2020}, pages 7871--7880. Association for Computational Linguistics.

\bibitem[{Li et~al.(2022)Li, Meng, Lin, Liu, Fu, Cao, Wang, and Zhou}]{alsocause}
Jiangnan Li, Fandong Meng, Zheng Lin, Rui Liu, Peng Fu, Yanan Cao, Weiping Wang, and Jie Zhou. 2022.
\newblock \href {https://doi.org/10.24963/IJCAI.2022/584} {Neutral utterances are also causes: Enhancing conversational causal emotion entailment with social commonsense knowledge}.
\newblock In \emph{Proceedings of the Thirty-First International Joint Conference on Artificial Intelligence, {IJCAI} 2022, Vienna, Austria, 23-29 July 2022}, pages 4209--4215. ijcai.org.

\bibitem[{Li et~al.(2016)Li, Galley, Brockett, Gao, and Dolan}]{Dist}
Jiwei Li, Michel Galley, Chris Brockett, Jianfeng Gao, and Bill Dolan. 2016.
\newblock \href {https://doi.org/10.18653/v1/n16-1014} {A diversity-promoting objective function for neural conversation models}.
\newblock In \emph{{NAACL} {HLT} 2016, The 2016 Conference of the North American Chapter of the Association for Computational Linguistics: Human Language Technologies, San Diego California, USA, June 12-17, 2016}, pages 110--119. The Association for Computational Linguistics.

\bibitem[{Li et~al.(2023)Li, Hessel, Yu, Ren, Chang, and Choi}]{Simbolic}
Liunian~Harold Li, Jack Hessel, Youngjae Yu, Xiang Ren, Kai{-}Wei Chang, and Yejin Choi. 2023.
\newblock \href {https://doi.org/10.18653/V1/2023.ACL-LONG.150} {Symbolic chain-of-thought distillation: Small models can also "think" step-by-step}.
\newblock In \emph{Proceedings of the 61st Annual Meeting of the Association for Computational Linguistics (Volume 1: Long Papers), {ACL} 2023, Toronto, Canada, July 9-14, 2023}, pages 2665--2679. Association for Computational Linguistics.

\bibitem[{Li et~al.(2020)Li, Li, Chen, and Ren}]{KEMP}
Qintong Li, Piji Li, Zhumin Chen, and Zhaochun Ren. 2020.
\newblock \href {https://arxiv.org/abs/2009.09708} {Empathetic dialogue generation via knowledge enhancing and emotion dependency modeling}.
\newblock \emph{CoRR}, abs/2009.09708.

\bibitem[{Lin(2004)}]{rouge}
Chin-Yew Lin. 2004.
\newblock Rouge: A package for automatic evaluation of summaries.
\newblock In \emph{Text summarization branches out}, pages 74--81.

\bibitem[{Lin et~al.(2019)Lin, Madotto, Shin, Xu, and Fung}]{MoEL}
Zhaojiang Lin, Andrea Madotto, Jamin Shin, Peng Xu, and Pascale Fung. 2019.
\newblock \href {https://doi.org/10.18653/v1/D19-1012} {{M}o{EL}: Mixture of empathetic listeners}.
\newblock In \emph{Proceedings of the 2019 Conference on Empirical Methods in Natural Language Processing and the 9th International Joint Conference on Natural Language Processing (EMNLP-IJCNLP)}, pages 121--132, Hong Kong, China. Association for Computational Linguistics.

\bibitem[{Liu et~al.(2022)Liu, Tan, Tao, Fu, Zhao, Liu, and Yan}]{ProphetChat}
Chang Liu, Xu~Tan, Chongyang Tao, Zhenxin Fu, Dongyan Zhao, Tie{-}Yan Liu, and Rui Yan. 2022.
\newblock \href {https://doi.org/10.18653/v1/2022.acl-long.68} {Prophetchat: Enhancing dialogue generation with simulation of future conversation}.
\newblock In \emph{Proceedings of the 60th Annual Meeting of the Association for Computational Linguistics (Volume 1: Long Papers), {ACL} 2022, Dublin, Ireland, May 22-27, 2022}, pages 962--973. Association for Computational Linguistics.

\bibitem[{Liu et~al.(2016)Liu, Lowe, Serban, Noseworthy, Charlin, and Pineau}]{HowNot}
Chia{-}Wei Liu, Ryan Lowe, Iulian Serban, Michael Noseworthy, Laurent Charlin, and Joelle Pineau. 2016.
\newblock \href {https://doi.org/10.18653/v1/d16-1230} {How {NOT} to evaluate your dialogue system: An empirical study of unsupervised evaluation metrics for dialogue response generation}.
\newblock In \emph{Proceedings of the 2016 Conference on Empirical Methods in Natural Language Processing, {EMNLP} 2016, Austin, Texas, USA, November 1-4, 2016}, pages 2122--2132. The Association for Computational Linguistics.

\bibitem[{Liu et~al.(2021)Liu, Zheng, Demasi, Sabour, Li, Yu, Jiang, and Huang}]{ESCdata}
Siyang Liu, Chujie Zheng, Orianna Demasi, Sahand Sabour, Yu~Li, Zhou Yu, Yong Jiang, and Minlie Huang. 2021.
\newblock \href {https://doi.org/10.18653/v1/2021.acl-long.269} {Towards emotional support dialog systems}.
\newblock In \emph{Proceedings of the 59th Annual Meeting of the Association for Computational Linguistics and the 11th International Joint Conference on Natural Language Processing, {ACL/IJCNLP} 2021, (Volume 1: Long Papers), Virtual Event, August 1-6, 2021}, pages 3469--3483. Association for Computational Linguistics.

\bibitem[{Liu et~al.(2023)Liu, Iter, Xu, Wang, Xu, and Zhu}]{G-EVAL}
Yang Liu, Dan Iter, Yichong Xu, Shuohang Wang, Ruochen Xu, and Chenguang Zhu. 2023.
\newblock \href {https://doi.org/10.48550/arXiv.2303.16634} {G-eval: {NLG} evaluation using {GPT-4} with better human alignment}.
\newblock \emph{CoRR}, abs/2303.16634.

\bibitem[{Majumder et~al.(2020)Majumder, Hong, Peng, Lu, Ghosal, Gelbukh, Mihalcea, and Poria}]{MIME}
Navonil Majumder, Pengfei Hong, Shanshan Peng, Jiankun Lu, Deepanway Ghosal, Alexander Gelbukh, Rada Mihalcea, and Soujanya Poria. 2020.
\newblock \href {https://doi.org/10.18653/v1/2020.emnlp-main.721} {{MIME}: {MIM}icking emotions for empathetic response generation}.
\newblock In \emph{Proceedings of the 2020 Conference on Empirical Methods in Natural Language Processing (EMNLP)}, pages 8968--8979, Online. Association for Computational Linguistics.

\bibitem[{MetaAI(2024)}]{Llama3.1}
MetaAI. 2024.
\newblock \href {https://ai.meta.com/blog/meta-llama-3-1/} {Introducing llama 3.1: Our most capable models to date}.
\newblock Accessed on July 23, 2024.

\bibitem[{OpenAI(2022)}]{ChatGPT}
OpenAI. 2022.
\newblock \href {https://openai.com/blog/chatgpt/} {Chatgpt: Optimizing language models for dialogue}.
\newblock Accessed on January 10, 2023.

\bibitem[{OpenAI(2023)}]{GPT4}
OpenAI. 2023.
\newblock \href {https://doi.org/10.48550/arXiv.2303.08774} {{GPT-4} technical report}.
\newblock \emph{CoRR}, abs/2303.08774.

\bibitem[{Papineni et~al.(2002)Papineni, Roukos, Ward, and Zhu}]{BLEU}
Kishore Papineni, Salim Roukos, Todd Ward, and Wei{-}Jing Zhu. 2002.
\newblock \href {https://doi.org/10.3115/1073083.1073135} {Bleu: a method for automatic evaluation of machine translation}.
\newblock In \emph{Proceedings of the 40th Annual Meeting of the Association for Computational Linguistics, July 6-12, 2002, Philadelphia, PA, {USA}}, pages 311--318. {ACL}.

\bibitem[{Peng et~al.(2022)Peng, Hu, Xing, Xie, Sun, and Li}]{GLHG}
Wei Peng, Yue Hu, Luxi Xing, Yuqiang Xie, Yajing Sun, and Yunpeng Li. 2022.
\newblock \href {https://doi.org/10.24963/IJCAI.2022/600} {Control globally, understand locally: {A} global-to-local hierarchical graph network for emotional support conversation}.
\newblock In \emph{Proceedings of the Thirty-First International Joint Conference on Artificial Intelligence, {IJCAI} 2022, Vienna, Austria, 23-29 July 2022}, pages 4324--4330. ijcai.org.

\bibitem[{Qian et~al.(2023)Qian, Zhang, and Liu}]{ED_ChatGPT}
Yushan Qian, Weinan Zhang, and Ting Liu. 2023.
\newblock \href {https://doi.org/10.18653/v1/2023.findings-emnlp.433} {Harnessing the power of large language models for empathetic response generation: Empirical investigations and improvements}.
\newblock In \emph{Findings of the Association for Computational Linguistics: EMNLP 2023}, pages 6516--6528, Singapore. Association for Computational Linguistics.

\bibitem[{Rashkin et~al.(2019)Rashkin, Smith, Li, and Boureau}]{EDdata}
Hannah Rashkin, Eric~Michael Smith, Margaret Li, and Y{-}Lan Boureau. 2019.
\newblock \href {https://doi.org/10.18653/v1/p19-1534} {Towards empathetic open-domain conversation models: {A} new benchmark and dataset}.
\newblock In \emph{Proceedings of the 57th Conference of the Association for Computational Linguistics, {ACL} 2019, Florence, Italy, July 28- August 2, 2019, Volume 1: Long Papers}, pages 5370--5381. Association for Computational Linguistics.

\bibitem[{Roller et~al.(2021)Roller, Dinan, Goyal, Ju, Williamson, Liu, Xu, Ott, Smith, Boureau, and Weston}]{Blenderbot}
Stephen Roller, Emily Dinan, Naman Goyal, Da~Ju, Mary Williamson, Yinhan Liu, Jing Xu, Myle Ott, Eric~Michael Smith, Y{-}Lan Boureau, and Jason Weston. 2021.
\newblock \href {https://doi.org/10.18653/V1/2021.EACL-MAIN.24} {Recipes for building an open-domain chatbot}.
\newblock In \emph{Proceedings of the 16th Conference of the European Chapter of the Association for Computational Linguistics: Main Volume, {EACL} 2021, Online, April 19 - 23, 2021}, pages 300--325. Association for Computational Linguistics.

\bibitem[{Sabour et~al.(2021)Sabour, Zheng, and Huang}]{CEM}
Sahand Sabour, Chujie Zheng, and Minlie Huang. 2021.
\newblock \href {https://arxiv.org/abs/2109.05739} {{CEM:} commonsense-aware empathetic response generation}.
\newblock \emph{CoRR}, abs/2109.05739.

\bibitem[{Santra et~al.(2023)Santra, Basak, De, Gupta, and Goyal}]{Frugal_Prompting}
Bishal Santra, Sakya Basak, Abhinandan De, Manish Gupta, and Pawan Goyal. 2023.
\newblock \href {https://aclanthology.org/2023.findings-emnlp.290} {Frugal prompting for dialog models}.
\newblock In \emph{Findings of the Association for Computational Linguistics: {EMNLP} 2023, Singapore, December 6-10, 2023}, pages 4383--4407. Association for Computational Linguistics.

\bibitem[{Shen et~al.(2022)Shen, Ghosal, Majumder, Lim, Mihalcea, and Poria}]{CICEROv2}
Siqi Shen, Deepanway Ghosal, Navonil Majumder, Henry Lim, Rada Mihalcea, and Soujanya Poria. 2022.
\newblock \href {https://doi.org/10.48550/arXiv.2210.02890} {Multiview contextual commonsense inference: {A} new dataset and task}.
\newblock \emph{CoRR}, abs/2210.02890.

\bibitem[{Speer et~al.(2016)Speer, Chin, and Havasi}]{ConceptNet}
Robyn Speer, Joshua Chin, and Catherine Havasi. 2016.
\newblock \href {https://arxiv.org/abs/1612.03975} {Conceptnet 5.5: An open multilingual graph of general knowledge}.
\newblock \emph{CoRR}, abs/1612.03975.

\bibitem[{Touvron et~al.(2023)Touvron, Martin, Stone, Albert, Almahairi, Babaei, Bashlykov, Batra, Bhargava, Bhosale, Bikel, Blecher, Canton{-}Ferrer, Chen, Cucurull, Esiobu, Fernandes, Fu, Fu, Fuller, Gao, Goswami, Goyal, Hartshorn, Hosseini, Hou, Inan, Kardas, Kerkez, Khabsa, Kloumann, Korenev, Koura, Lachaux, Lavril, Lee, Liskovich, Lu, Mao, Martinet, Mihaylov, Mishra, Molybog, Nie, Poulton, Reizenstein, Rungta, Saladi, Schelten, Silva, Smith, Subramanian, Tan, Tang, Taylor, Williams, Kuan, Xu, Yan, Zarov, Zhang, Fan, Kambadur, Narang, Rodriguez, Stojnic, Edunov, and Scialom}]{Llama2}
Hugo Touvron, Louis Martin, Kevin Stone, Peter Albert, Amjad Almahairi, Yasmine Babaei, Nikolay Bashlykov, Soumya Batra, Prajjwal Bhargava, Shruti Bhosale, Dan Bikel, Lukas Blecher, Cristian Canton{-}Ferrer, Moya Chen, Guillem Cucurull, David Esiobu, Jude Fernandes, Jeremy Fu, Wenyin Fu, Brian Fuller, Cynthia Gao, Vedanuj Goswami, Naman Goyal, Anthony Hartshorn, Saghar Hosseini, Rui Hou, Hakan Inan, Marcin Kardas, Viktor Kerkez, Madian Khabsa, Isabel Kloumann, Artem Korenev, Punit~Singh Koura, Marie{-}Anne Lachaux, Thibaut Lavril, Jenya Lee, Diana Liskovich, Yinghai Lu, Yuning Mao, Xavier Martinet, Todor Mihaylov, Pushkar Mishra, Igor Molybog, Yixin Nie, Andrew Poulton, Jeremy Reizenstein, Rashi Rungta, Kalyan Saladi, Alan Schelten, Ruan Silva, Eric~Michael Smith, Ranjan Subramanian, Xiaoqing~Ellen Tan, Binh Tang, Ross Taylor, Adina Williams, Jian~Xiang Kuan, Puxin Xu, Zheng Yan, Iliyan Zarov, Yuchen Zhang, Angela Fan, Melanie Kambadur, Sharan Narang, Aur{\'{e}}lien Rodriguez, Robert Stojnic, Sergey Edunov,
  and Thomas Scialom. 2023.
\newblock \href {https://doi.org/10.48550/ARXIV.2307.09288} {Llama 2: Open foundation and fine-tuned chat models}.
\newblock \emph{CoRR}, abs/2307.09288.

\bibitem[{Tu et~al.(2022)Tu, Li, Cui, Wang, Wen, and Yan}]{MISC}
Quan Tu, Yanran Li, Jianwei Cui, Bin Wang, Ji{-}Rong Wen, and Rui Yan. 2022.
\newblock \href {https://doi.org/10.18653/V1/2022.ACL-LONG.25} {{MISC:} {A} mixed strategy-aware model integrating {COMET} for emotional support conversation}.
\newblock In \emph{Proceedings of the 60th Annual Meeting of the Association for Computational Linguistics (Volume 1: Long Papers), {ACL} 2022, Dublin, Ireland, May 22-27, 2022}, pages 308--319. Association for Computational Linguistics.

\bibitem[{Vaswani et~al.(2017)Vaswani, Shazeer, Parmar, Uszkoreit, Jones, Gomez, Kaiser, and Polosukhin}]{TRANSFORMER}
Ashish Vaswani, Noam Shazeer, Niki Parmar, Jakob Uszkoreit, Llion Jones, Aidan~N. Gomez, Lukasz Kaiser, and Illia Polosukhin. 2017.
\newblock \href {https://proceedings.neurips.cc/paper/2017/hash/3f5ee243547dee91fbd053c1c4a845aa-Abstract.html} {Attention is all you need}.
\newblock In \emph{Advances in Neural Information Processing Systems 30: Annual Conference on Neural Information Processing Systems 2017, December 4-9, 2017, Long Beach, CA, {USA}}, pages 5998--6008.

\bibitem[{Vedantam et~al.(2015)Vedantam, Zitnick, and Parikh}]{CIDEr}
Ramakrishna Vedantam, C.~Lawrence Zitnick, and Devi Parikh. 2015.
\newblock \href {https://doi.org/10.1109/CVPR.2015.7299087} {Cider: Consensus-based image description evaluation}.
\newblock In \emph{{IEEE} Conference on Computer Vision and Pattern Recognition, {CVPR} 2015, Boston, MA, USA, June 7-12, 2015}, pages 4566--4575. {IEEE} Computer Society.

\bibitem[{Wang et~al.(2023)Wang, Wang, Mi, Wang, Xu, and Wong}]{Cue-CoT}
Hongru Wang, Rui Wang, Fei Mi, Zezhong Wang, Ruifeng Xu, and Kam{-}Fai Wong. 2023.
\newblock \href {https://doi.org/10.48550/ARXIV.2305.11792} {Chain-of-thought prompting for responding to in-depth dialogue questions with {LLM}}.
\newblock \emph{CoRR}, abs/2305.11792.

\bibitem[{Wang et~al.(2022)Wang, Li, Lin, Meng, Yang, Wang, and Zhou}]{SEEK}
Lanrui Wang, Jiangnan Li, Zheng Lin, Fandong Meng, Chenxu Yang, Weiping Wang, and Jie Zhou. 2022.
\newblock \href {https://doi.org/10.48550/arXiv.2210.11715} {Empathetic dialogue generation via sensitive emotion recognition and sensible knowledge selection}.
\newblock \emph{CoRR}, abs/2210.11715.

\bibitem[{Welivita and Pu(2020)}]{Taxonomy}
Anuradha Welivita and Pearl Pu. 2020.
\newblock \href {https://doi.org/10.18653/v1/2020.coling-main.429} {A taxonomy of empathetic response intents in human social conversations}.
\newblock In \emph{Proceedings of the 28th International Conference on Computational Linguistics, {COLING} 2020, Barcelona, Spain (Online), December 8-13, 2020}, pages 4886--4899. International Committee on Computational Linguistics.

\bibitem[{Wolf et~al.(2020)Wolf, Debut, Sanh, Chaumond, Delangue, Moi, Cistac, Rault, Louf, Funtowicz, Davison, Shleifer, von Platen, Ma, Jernite, Plu, Xu, Scao, Gugger, Drame, Lhoest, and Rush}]{Transformers}
Thomas Wolf, Lysandre Debut, Victor Sanh, Julien Chaumond, Clement Delangue, Anthony Moi, Pierric Cistac, Tim Rault, R{\'{e}}mi Louf, Morgan Funtowicz, Joe Davison, Sam Shleifer, Patrick von Platen, Clara Ma, Yacine Jernite, Julien Plu, Canwen Xu, Teven~Le Scao, Sylvain Gugger, Mariama Drame, Quentin Lhoest, and Alexander~M. Rush. 2020.
\newblock \href {https://doi.org/10.18653/v1/2020.emnlp-demos.6} {Transformers: State-of-the-art natural language processing}.
\newblock In \emph{Proceedings of the 2020 Conference on Empirical Methods in Natural Language Processing: System Demonstrations, {EMNLP} 2020 - Demos, Online, November 16-20, 2020}, pages 38--45. Association for Computational Linguistics.

\bibitem[{Yang et~al.(2022)Yang, Lin, Li, Meng, Wang, Wang, and Zhou}]{TAKE}
Chenxu Yang, Zheng Lin, Jiangnan Li, Fandong Meng, Weiping Wang, Lanrui Wang, and Jie Zhou. 2022.
\newblock \href {https://aclanthology.org/2022.coling-1.20} {{TAKE:} topic-shift aware knowledge selection for dialogue generation}.
\newblock In \emph{Proceedings of the 29th International Conference on Computational Linguistics, {COLING} 2022, Gyeongju, Republic of Korea, October 12-17, 2022}, pages 253--265. International Committee on Computational Linguistics.

\bibitem[{Zhao et~al.(2023{\natexlab{a}})Zhao, Zhao, Lu, and Qin}]{EmpSoA}
Weixiang Zhao, Yanyan Zhao, Xin Lu, and Bing Qin. 2023{\natexlab{a}}.
\newblock \href {https://doi.org/10.18653/V1/2023.FINDINGS-ACL.843} {Don't lose yourself! empathetic response generation via explicit self-other awareness}.
\newblock In \emph{Findings of the Association for Computational Linguistics: {ACL} 2023, Toronto, Canada, July 9-14, 2023}, pages 13331--13344. Association for Computational Linguistics.

\bibitem[{Zhao et~al.(2023{\natexlab{b}})Zhao, Zhao, Lu, Wang, Tong, and Qin}]{IsChatGPT}
Weixiang Zhao, Yanyan Zhao, Xin Lu, Shilong Wang, Yanpeng Tong, and Bing Qin. 2023{\natexlab{b}}.
\newblock \href {https://doi.org/10.48550/ARXIV.2304.09582} {Is chatgpt equipped with emotional dialogue capabilities?}
\newblock \emph{CoRR}, abs/2304.09582.

\bibitem[{Zhou et~al.(2018)Zhou, Huang, Zhang, Zhu, and Liu}]{ECM}
Hao Zhou, Minlie Huang, Tianyang Zhang, Xiaoyan Zhu, and Bing Liu. 2018.
\newblock \href {https://www.aaai.org/ocs/index.php/AAAI/AAAI18/paper/view/16455} {Emotional chatting machine: Emotional conversation generation with internal and external memory}.
\newblock In \emph{Proceedings of the Thirty-Second {AAAI} Conference on Artificial Intelligence, (AAAI-18), the 30th innovative Applications of Artificial Intelligence (IAAI-18), and the 8th {AAAI} Symposium on Educational Advances in Artificial Intelligence (EAAI-18), New Orleans, Louisiana, USA, February 2-7, 2018}, pages 730--739. {AAAI} Press.

\bibitem[{Zhou et~al.(2023)Zhou, Zheng, Wang, Zhang, and Huang}]{CASE}
Jinfeng Zhou, Chujie Zheng, Bo~Wang, Zheng Zhang, and Minlie Huang. 2023.
\newblock \href {https://doi.org/10.18653/V1/2023.ACL-LONG.457} {{CASE:} aligning coarse-to-fine cognition and affection for empathetic response generation}.
\newblock In \emph{Proceedings of the 61st Annual Meeting of the Association for Computational Linguistics (Volume 1: Long Papers), {ACL} 2023, Toronto, Canada, July 9-14, 2023}, pages 8223--8237. Association for Computational Linguistics.

\bibitem[{Zhou et~al.(2022)Zhou, Cho, Jandaghi, Lee, Lin, Pujara, and Ren}]{Reflect}
Pei Zhou, Hyundong Cho, Pegah Jandaghi, Dong{-}Ho Lee, Bill~Yuchen Lin, Jay Pujara, and Xiang Ren. 2022.
\newblock \href {https://doi.org/10.18653/V1/2022.EMNLP-MAIN.714} {Reflect, not reflex: Inference-based common ground improves dialogue response quality}.
\newblock In \emph{Proceedings of the 2022 Conference on Empirical Methods in Natural Language Processing, {EMNLP} 2022, Abu Dhabi, United Arab Emirates, December 7-11, 2022}, pages 10450--10468. Association for Computational Linguistics.

\end{thebibliography}
\newpage

\appendix
\section{Four Categories of Commonsense Knowledge}
\label{appendix_cateies_cmk}
We mainly employ four categories of commonsense knowledge of our proposed paradigm, which is as follows.

\textbf{Cause} \quad \textit{What is the cause of the assistant to post the last utterance?}  We emphasize the crucial role of causality within the dialogue context. Similar to the approach outlined by \citet{CICEROv2} and previous investigations \citep{alsocause, MultiESC}, we delve into potential words or phrases that could lead to the desired response. 

\textbf{Subsequent Event} \quad \textit{What will be the potential subsequent events involving the assistant that may occur after the user's last utterance? } Conversations demonstrate a causal connection between past utterances to the ensuing responses. Dialogues contain a cause-and-effect connection between the context and the target response. Following \citep{CICEROv1}, we employ a language model to project potential scenarios that follow the dialogue history, which is a key factor in determining the assistant's response.

\textbf{Emotion reaction} \quad \textit{What is the emotional reaction of the user in their last utterance?} Emotion is a fundamental element in human conversation \citep{ECM}, acting as a natural means for individuals to express their feelings during dialogues. With explicit emotion traits, it is easier for chatbots to grasp a more profound understanding of the dialogue and anticipate the potential emotional content within the target response.


\definecolor{awesome}{rgb}{1.0, 0.13, 0.32}
\definecolor{azure(colorwheel)}{rgb}{0.0, 0.5, 1.0}
\definecolor{aureolin}{rgb}{0.99, 0.93, 0.0}
\definecolor{amber}{rgb}{0.99, 0.93, 0.0}
\definecolor{frenchrose}{rgb}{0.96, 0.29, 0.54}
\definecolor{coquelicot}{rgb}{1.0, 0.22, 0.0}
\definecolor{aliceblue}{rgb}{0.9, 0.9, 0.9}

\lstdefinelanguage{prompt}{
    frame=shadowbox,
    framerule=0.5pt,
    framesep=2pt,
    breaklines=true,
    backgroundcolor=\color{aliceblue},
    basicstyle=\fontsize{9pt}{9pt}\selectfont\ttfamily,
    commentstyle=\color{cyan},
    morecomment=[l]{//},
    moredelim=[is][\color{frenchrose}\bfseries]{<<<}{>>>},
    moredelim=[is][\color{awesome}\bfseries]{***}{***},
    moredelim=[is][\color{azure(colorwheel)}\bfseries]{///}{///},
    moredelim=[is][\color{coquelicot}\bfseries]{|||}{|||},
}

\lstdefinestyle{mystyle}{
    basicstyle=\fontsize{9pt}{9pt}\selectfont\ttfamily,
    breakatwhitespace=false,
    backgroundcolor=\color{aliceblue},
    xleftmargin=1pt,
    breaklines=true,
    breakindent=0pt,
}
\lstset{style=mystyle}

The template input for prompting Large Language Models generating prophetic commonsense inference is as follows: \\
\begin{figure*}[!htb]
\begin{lstlisting}[language=prompt]
***[SYSTEM]***
Given a dyadic dialogue clip between a listener and a speaker, the objective is to comprehend the dialogue and make inferences to identify the underlying cause of the latest utterance stated by the listener (the reason contributing to the utterance stated by the listener). 

I will provide an example of a conversation clip and an explanation of the causes, which are as follows:

///(1)Speaker: Job interviews always make me sweat bullets, makes me uncomfortable in general to be looked at under a microscope like that.
(2)Listener: Don't be nervous. Just be prepared.
(3)Speaker: I feel like getting prepared and then having a curve ball thrown at you throws you off.
(4)Listener: Yes but if you stay calm it will be ok.///

What is the cause of the listener to post the next response? Please make inferences based on the utterances before the last utterance of the conversation. Please generate the answer like this: Answer: The cause of the listener's last utterance is to reassure and encourage the speaker, emphasizing the importance of staying calm despite unexpected challenges during a job interview.

***[USER]***
Now, generate one concise and relevant inference (no more than 40 words) of the cause of the last utterance. The conversation clip is: 

|||{context}|||

What is the cause of the listener to post the next response?

Answer:
\end{lstlisting}
\caption{Prompt template for Visionary Commonsense acquisition.}
\label{figure_visionary_commonsense_acquisition}
\end{figure*} 

\textbf{Intention} \quad \textit{What is the assistant's intent to post the last utterance according to the emotional reaction of the user?} Dialogue intention is a focal point in the realm of dialogue generation \citep{Taxonomy}. It comprises the underlying logic and objectives guiding the forthcoming conversation, thus forming a vital aspect in contextual understanding and response generation. 

The above four categories of commonsense inference are all used in our paradigm, acting as intermediate reasoning steps for steering language models for better dialogue comprehension and more empathetic responses.



\begin{figure*}[!htb]
\begin{lstlisting}[language=prompt]
***[SYSTEM]***
Assuming that you are a highly empathetic person, there is a dyadic dialogue clip between a listener and a speaker. You should first identify emotion of the speaker in the dyadic dialogue clip, and then generate a concise, relevant, and empathetic response for the following conversation.
Please generate a response that incorporates relevant common-sense knowledge: 

The underlying cause of the listener's next utterance (the reason contributing to response) is: |||{cause}|||.

The subsequent event about the listener that happens or could happen following the last utterance stated by the listener: |||{subsequent}|||.

The possible emotional reaction of the speaker in response to the last utterance stated by the speaker is: |||{emotion}|||.

The listener's intent to post the last utterance according to the emotion reaction of the speaker is: |||{intent}|||.

***[USER]***
|||{speaker_utterance1}|||

***[Assistant]***
|||{listener_utterance1}|||

***[USER]***
|||{speaker_utterance2}|||

...
\end{lstlisting}
\caption{Prompt template for \textit{Sibyl} training.}
\label{figure_sibyl_training}
\end{figure*}

\section{Details of Baseline Methods} \label{sec:appendix_baselines}
In this Section, we present the details of the baseline methods we compared with our proposed paradigm \textit{Sibyl}. \\
\textbf{CASE} \citep{CASE}: A model trained from scratch with \textbf{vanilla transformers} \citep{TRANSFORMER} on ED dataset. This work utilizes COMET \citep{COMET} and ConceptNet \citep{ConceptNet} as auxiliary information and constructs a conditional graph to represent all plausible causalities between the user's emotions and experience. We select CASE as a reference of the traditional from-scratched method on the ED benchmark. \\
\textbf{M-Cue CoT} \citep{Cue-CoT}: A multi-step prompting mechanism to trace the status of users during the conversation, performing complex reasoning and planning before generating the final response. As \textbf{M-Cue CoT} is a prompting engineering method, we only compare it under prompted-based experiments settings. \\
\textbf{LLaMA} and Flan-t5-xl \citep{llama3,flant5}: To evaluate the performance of basic open-source foundation models, we utilize LLaMA and Flan-t5-xl as the primary backbone models. The experimental results reveal that the finetuned generator responds solely based on the context of the dialogue. \\
\textbf{+ COMET} \citep{COMET}: A Seq2seq model enhanced by external knowledge comes from ATOMIC \citep{ATOMIC} which makes inferences based on the last utterance of context. Following \citep{CEM, SEEK}, we apply COMET to generate five types of commonsense knowledge: the effect of the person (xEffect), the reaction of the person speaking the corresponding sentence (xReact), the intent before the person speaking (xIntent), what the person needs (xNeed), and what the person wants after speaking the sentence (xWant). \\
\textbf{+ DOCTOR} \citep{DialogueCoT}: 
As a dialogue Chain-of-Thought commonsense reasoner, DOCTOR adeptly integrates implicit information from dialogues to formulate responses. It is trained on various Open-Domain dialogue datasets, equipping it with a strong ability to generalize across Out-of-Domain data and reliably evaluate the psychological states and topics of discussion between interlocutors.\\
\textbf{+ DIALeCT} \citep{CICEROv2}: DIALeCT is a specialized model focused on dialogue-based commonsense knowledge. It is trained across a broad spectrum of dialogue-related tasks and open-domain dialogue datasets. This model is proficient in harnessing structural information from the entire dialogue context, rather than merely concentrating on specific utterances.

\section{Details of LLMs-based evaluation} \label{sec:appendix_GEVal}

The absence of labor-free and practical evaluation metrics has been a persistent challenge within the field of NLP research. Thanks to the rise of LLMs, several studies have explored the utilization of LLMs in assessing content generated by neural models. \cite{GPTscore} propose a direct approach, using LLMs as reference-free evaluators for Natural Language Generation (NLG), viewing the evaluation process as a probability calculation. Moreover, \cite{G-EVAL} and \cite{closergeval} introduce a prompt-based framework for LLMs, ensuring adherence to the generated instructions and offering a more detailed continuous score by adjusting the discrete scores based on their token probabilities.

We apply G-Eval \citep{G-EVAL, closergeval} to assess the Naturalness (\textbf{Nat.}) and Coherence (\textbf{Coh.}) of responses from baseline models that utilize commonsense knowledge in diverse ways. For task-specific requirements, we compare Empathy (\textbf{Emp.}) in the context of \textsc{EmpatheticDialogues} and Supportiveness (\textbf{Sup.}) for ESConv. As the token probabilities of GPT-4o are unavailable, we set '$n = 20, temperature = 1, top_p = 1$' to sample 20 times to estimate the token probabilities. 

Strictly following the rating strategy \citep{G-EVAL}, we prompt  \textit{gpt-4o} to discretely rate 1 to 3 points to these generated responses. Specifically, we require the LLMs to rate 1 when the generated response fails to meet a certain aspect. Rating a '2-point' means the response is totally ok, and meets the certain requirement to some extent. For responses that actually meet the desired demands, LLM is asked to give a '3-point' rating.


We randomly selected 200 data from both ED and ESConv datasets to perform G-Eval evaluation. Calculating the average weighted score of sampled data, the comparison result is shown in Table \ref{tab: G_eval_SFT} and Table \ref{tab: G_eval_ICL}, \textit{Sibyl} outperforms all strong baseline of commonsense inference in all aspects. The prompt template is specified in Figure \ref{Geval_template}.
\begin{table}[]
\centering
\scalebox{0.76}{
\begin{tabular}{l|ccc|ccc}
\toprule
\textbf{}      & \multicolumn{3}{c|}{\textbf{ED}}                 & \multicolumn{3}{c}{\textbf{ESConv}} \\ \cline{2-7} 
\multicolumn{1}{l|}{} & \textbf{Nat.} & \textbf{Emp.} & \textbf{Coh.} & \textbf{Nat.} & \textbf{Sup.} & \textbf{Coh.} \\ \hline
CASE        & 2.053 & 1.539	& 1.995 & -     & -     & -    \\
MultiESC    &  -    &  -    &   -   & 2.092 & 1.23  & 1.812    \\
LLaMA3.1      & 2.512 & 1.849 & 2.635 & 2.332 & 1.376 & 2.214  \\
+ COMET     & 2.464 & 1.747 & 2.646 & 2.368 & 1.944 & 2.465  \\
+ DOCTOR    & 2.503 & 2.088 & 2.653 & 2.349 & 1.408 & 2.496  \\
+ DIALeCT   & 2.441 & 1.115 & 2.644 & 2.381 & 1.867 & 2.526  \\
+ \textit{Sibyl}  & \textbf{2.568} & \textbf{2.396} & \textbf{2.774} & \textbf{2.387} & \textbf{1.958} & \textbf{2.599}  \\ \bottomrule
\end{tabular}
}
\caption{LLMs based Evaluation results on \textsc{EpatheticDialogues} (ED)  and ESConv dataset under Supervised Finetyuning.}
\label{tab: G_eval_SFT}
\end{table}

\begin{table}[]
\centering
\scalebox{0.72}{
\begin{tabular}{l|ccc|ccc}
\toprule
\multicolumn{1}{l|}{} & \multicolumn{3}{c|}{\textbf{ED}}                  & \multicolumn{3}{c}{\textbf{ESConv}}        \\ \cline{2-7} 
\multicolumn{1}{l|}{} & \textbf{Nat.} & \textbf{Emp.} & \textbf{Coh.} & \textbf{Nat.} & \textbf{Sup.} & \textbf{Coh.} \\ \hline
GPT-4    & 2.19           &  2.171          & \textbf{2.192} & 1.838          & 1.983 & 1.713 \\
+ COMET        & 2.188          &  2.176        & 2.188 & 1.842 & 1.979 & 1.712          \\
+ DIALeCT& 2.126 & 1.793 & 2.186 & 1.841 & 1.793 & 1.71 \\
+ M-Cue CoT & 2.189 & 1.792 & 2.124 & 1.841 & 1.982 & 1.716 \\
+ \textit{Sibyl}         & \textbf{2.191} &  \textbf{2.176} & 2.191    & \textbf{1.846 }   & \textbf{1.984} & \textbf{1.717}          \\ \bottomrule
\end{tabular}
}
\caption{LLMs based Evaluation results on \textsc{EpatheticDialogues} (ED) and ESConv dataset under In-Context Learning.}
\label{tab: G_eval_ICL}
\end{table}

\begin{table*}[]
\centering
\scalebox{0.78}{
\begin{tabular}{lccccccc}
\toprule
\textbf{Models} & \textbf{BLEU-1/2/3/4} & \textbf{Dist-1/2/3} & \textbf{ROU\_L.} & \textbf{MET.} & \textbf{Ave.} & \textbf{Ext.} & \textbf{CIDEr} \\ \hline
Normal Transformer & 13.74/5.70/2.85/1.55 & 0.66/3.16/6.40 & 16.12 & 12.02 & \textbf{86.82} & 51.87 & 15.04  \\
EmpSOA & 16.16/7.65/4.29/\textbf{2.71} & 0.74/3.67/7.58 & 17.69 & 13.51 & 85.83 & 51.52 & 19.2 \\
CASE & 15.99/7.41/3.90/2.29 & 0.64/3.02/5.98 & 18 & 7.77 & 87 & 51.02 & 18.12 \\
Normal Transformer + \textit{Sibyl} & \textbf{18.08/8.48/4.47}/2.57 & \textbf{0.87/5.76/14.20} & \textbf{18.86} & \textbf{14.92} & \textbf{86.82} & \textbf{52.38} & \textbf{22.41} \\ \bottomrule
\end{tabular}
}
\caption{Automatic evaluation results for models with transformers trained from scratch (Normal Transformer) on the ED dataset. The best outcomes are highlighted in \textbf{bold}.}
\label{tab: ED_backbone}
\end{table*}
\begin{table*}[]
\centering
\scalebox{0.78}{
\begin{tabular}{lcccccc}
\toprule
 \textbf{Models} & \textbf{BLEU-1/2/3/4} & \textbf{ROU\_L.} & \textbf{MET.} & \textbf{Ave.} & \textbf{Ext.} & \textbf{CIDEr} \\ \hline
 Normal TRS & 15.61/4.83/1.88/0.80 & 13.59 & 6.13 & 88.45 & 45.11 & 4.80 \\
 Normal TRS + \textit{Sibyl} & \textbf{21.76/11.16/4.64/1.87}  & \textbf{22.16} & \textbf{21.72} & \textbf{90.41} & \textbf{52.01} & \textbf{5.56} \\ \hline
 Blenderbot-Joint & 17.62/6.91/2.81/1.66 & 17.94 & 7.54 & - & - & -\\
 MISC &  - / 7.31 / - / 2.20 & 17.91 & - & - & 	- & - \\
Blenderbot + \textit{Sibyl} & \textbf{19.08/7.41/3.4/2.27} & \textbf{17.97} & - & 91.11 & 	50.45 & 7.5 \\ \hline
MultiESC (Bart-base)  &  18.73/7.58/3.93/2.23 & 18.57 & 8.35 & 91.14 & 48.18 & 22.23 \\
MultiESC + \textit{Sibyl} & \textbf{19.78/7.98/4.12/2.38} & \textbf{18.6} & \textbf{8.54} & 91.03 & 46.85 & \textbf{23.1} \\ \bottomrule
\end{tabular}
}
\caption{Automatic evaluation results for models with transformers trained from scratch (Normal Transformer), Blender-small-90M, and Bart-base on the ESConv dataset. }
\label{tab: ESC_backbone}
\end{table*}

\begin{table*}[]
\centering
\scalebox{0.75}{
\begin{tabular}{@{}m{3cm}m{17cm}@{}}
\toprule
\textbf{}    & \textit{Speaker:} Hi, I feel so lonely sometimes because all my friends live in a different country. \\ \cmidrule(l){2-2} 
 & \textit{Listener:} Oh, I'm sure you are lonely. Maybe you can join some kind of club that lets you meet new friends?       \\ \cmidrule(l){2-2} 
\textbf{Dialogue Context}      & \textit{Speaker:} I was thinking about it! I wanted to join a group for local moms. \\ \cmidrule(l){2-2} 
 & \textbf{Response}: That's a good idea! This way you can also meet friends for yourself, but also maybe meet new friends for your children to hang out with while you do with their moms! \\ \midrule
\textbf{+ COMET}       &  That would be fab!  Do you live in a big city or a small town?    \\
 \textbf{+ DOCTOR} & The speaker wants to join a club that allows them to meet new friends. The listener suggests that the speaker join a local club that would be suitable for their area. \\ 
\textbf{+ DIALeCT}       & That would be great, I'm sure you will have great luck! \\ \midrule
  & \textbf{Subsequent events}:  The listener is likely to suggest specific activities or events that the speaker can {\color[HTML]{FF0000} participate in to meet new friends}, showing a proactive and helpful approach to the conversation.  \\ \cmidrule(l){2-2} 
 \multirow{2}{*}{\textbf{Visionary}} & \textbf{Emotion state}: The speaker feels hopeful and appreciates the listener's suggestion to join a group for local moms, as it aligns with their desire to meet new friends.\\ \cmidrule(l){2-2} 
  \multirow{-3}{*}{\textbf{Commonsense}} & \textbf{Cause}: The listener is motivated by empathy and the desire to offer practical solutions, encouraging the speaker to pursue {\color[HTML]{FF0000}social connections} .\\ \cmidrule(l){2-2} 
 & \textbf{Intent}: To encourage the speaker, acknowledging the potential benefits of joining a group for local moms and expressing hope that it will lead to {\color[HTML]{FF0000} positive outcomes for both the speaker and their children}. \\  \midrule
\textbf{+ \textit{Sibyl} (Ours) }          & That would be a great idea. You can {\color[HTML]{FF0000} make friends} for {\color[HTML]{FF0000} yourself and for your children}.   \\ \bottomrule
\end{tabular}
}
\caption{ An example involving responses from different versions of LLaMA models which are enhanced with different commonsense knowledge. The words relating to commonsense knowledge are highlighted in {\color[HTML]{FF0000} red}, while phrases in red signify the connection with knowledge and dialogue history.}
\label{tab: casestudy}
\end{table*}

\begin{figure*}[!htb]
\begin{lstlisting}[language=prompt]
***[SYSTEM]***
Your task is to rate the responses on one metric.
Please make sure you read and understand these instructions carefully. Please keep this conversation history open while reviewing, and refer to it as needed.
Evaluation Criteria:
Empathy (1-3) Is the response empathetically written?

- A score of 1 (bad) means that the response is not empathetic.
- A score of 2 (ok) means the response is totally ok, but empathetic to some extent.
- A score of 3 (good) means the response is empathetic, showing the Listener understands the User's emotional state and situation.
Evaluation Steps:
///1. Read the conversation, the conversation between the two individuals.
2. Read the potential response for the next turn in the conversation.
3. Evaluate the response based on its Empathy, using the provided criteria.
4. Assign a rating score of 1, 2, or 3 based on the evaluation.///

***[USER]***
Conversation History:
|||{Dialogue History}|||
Response:
|||{Response}|||

Evaluation Form (Answer by starting with "Analysis:" to analyze the given example regarding the evaluation criteria as concise as possible, and then give the numeric rating on the next line by "Rating:"):
Empathy:
\end{lstlisting}
\caption{Prompt template for evaluating the empathy of the generated response using GPT-4.}
\label{Geval_template}
\end{figure*}

\end{document}